\documentclass[10pt,twocolumn,letterpaper]{article}
\usepackage{cvpr}    

\usepackage[accsupp]{axessibility}
\definecolor{cvprblue}{rgb}{0.21,0.49,0.74}
\usepackage[pagebackref,breaklinks,colorlinks,citecolor=cvprblue]{hyperref}

\def\paperID{14931} 

\def\confName{CVPR}
\def\confYear{2025}
\usepackage{algorithm}
\usepackage{algorithmic}
\usepackage{amsmath}
\usepackage{amssymb}
\usepackage{amsfonts}
\usepackage{amsthm}
\newtheorem{theorem}{Theorem}

\newtheorem{definition}{Definition}
\usepackage{multirow}
\usepackage{threeparttable}

\usepackage{hhline}

\usepackage{xcolor}
\usepackage{lineno}

\definecolor{deepblue}{rgb}{0.0, 0.0, 0.85}

\title{Tightening Robustness Verification of MaxPool-based Neural Networks via Minimizing the Over-Approximation Zone}

\author{
    Yuan Xiao$^{1,3}$,Yuchen Chen$^{1}$, Shiqing Ma$^{2}$,  Chunrong Fang$^{1,3}$\thanks{Chunrong Fang and Zhenyu Chen are the corresponding authors.}, Tongtong Bai$^{1}$, Mingzheng Gu$^{1}$, \\Yuxin Cheng$^{1}$, Yanwei Chen$^{1}$, Zhenyu Chen$^{1, 3*}$\\
    $^{1}$ State Key Laboratory for Novel Software Technology, Nanjing University, China \\ 
    $^{2}$University of Massachusetts Amherst, United States
      \\ $^{3}$ Shenzhen Research Institute, Nanjing University, China
}

\begin{document}
\maketitle              

\begin{abstract}
    The robustness of neural network classifiers is important in the safety-critical domain and can be quantified by robustness verification. At present, efficient and scalable verification techniques are always sound but incomplete, and thus, the improvement of verified robustness results is the key criterion to evaluate the performance of incomplete verification approaches. The multi-variate function MaxPool is widely adopted yet challenging to verify.
    In this paper, we present \textbf{Ti-Lin}, a robustness verifier for MaxPool-based CNNs with \textbf{Ti}ght \textbf{Lin}ear Approximation. Following the sequel of minimizing the over-approximation zone of the non-linear function of CNNs, we are the first to propose the provably neuron-wise tightest linear bounds for the MaxPool function. By our proposed linear bounds, we can certify larger robustness results for CNNs. 
    We evaluate the effectiveness of Ti-Lin on different verification frameworks with open-sourced benchmarks, including LeNet, PointNet, and networks trained on the MNIST, CIFAR-10, Tiny ImageNet and ModelNet40 datasets. Experimental results show that Ti-Lin significantly outperforms the state-of-the-art methods across all networks with up to 78.6\% improvement in terms of the certified accuracy with almost the same time consumption as the fastest tool. 
    Our code is available at \url{https://github.com/xiaoyuanpigo/Ti-Lin-Hybrid-Lin}.
\end{abstract}

\section{Introduction}
\label{section1}
Although neural networks achieve remarkable success in many difficult classification tasks, researchers discover non-robust neural networks are vulnerable to perturbation from the environment and adversarial attacks~\cite{DBLP:conf/issta/00020LJMZ22,moosavi2017universal,szegedy2013intriguing,DBLP:conf/issta/ZhaoCWYS021}. 
In some safety-critical domains, such as autonomous driving~\cite{gopinath2018deepsafe} and face recognition~\cite{goswami2018unravelling}, some subtle adversarial perturbation is extremely imperceptible and harmful and may cause disastrous consequences. 
Therefore, there is an urgent need to certify model robustness guarantees against adversarial attacks, which can provide certified defense against any possible attacks~\cite{boopathy2019cnn,du2021cert,singh2018fast}. Mathematically, verifiers aim to prove or disprove the correctness of the output neurons in response to input values with minor perturbations.

The methodology of robustness verification can be divided into complete verifiers and incomplete verifiers.
As complete verification is NP-complete even for the simple ReLU-based fully-connected networks~\cite{katz2017reluplex}, it is necessary to risk some verification precision loss to improve efficiency~\cite{fazlyab2019probabilistic,jia2020certified,weng2019proven}. 
State-of-the-art verifiers use advanced branching~\cite{muller2022prima,wang2021beta,ferrari2022complete} and relaxation techniques~\cite{singh2019abstract,henriksen2020efficient,imagestar,nnv} to reduce the precision loss while maintaining efficiency and scalability. Linear approximation is an efficient relaxation technique that can be used to handle complex functions beyond univariate functions, such as $Sigmoid(x)Tanh(y),x\cdot Sigmoid(y)$~\cite{ko2019popqorn,prover,du2021cert} and MaxPool~\cite{osip,boopathy2019cnn,singh2019abstract,lorenz2021robustness}. Furthermore, the flexibility of this technique allows it to be combined with other incomplete or complete verification algorithms~\cite{xu2020fast,wang2021beta,shi2023formal}. Linear approximation, in essence, provides a pair of upper and lower linear constraints $u(\boldsymbol{x}),l(\boldsymbol{x})$ to bound the output of the network's non-linear layer $f(\boldsymbol{x})$ according to the layer's input region $\{\boldsymbol{x}|\boldsymbol{x}\in[\boldsymbol{l},\boldsymbol{u}]\}$. Considerable efforts have been devoted to tightening linear approximation for various non-linear functions~\cite{lorenz2021robustness,wu2021tightening,lyu2020fastened,henriksen2020efficient,prover,xu2020fast}, which have proven successful in certifying more precise results. 
Some of these works propose the linear bounds that can produce the smallest over-approximation zone, a.k.a., the neuron-wise tightest linear bounds~\cite{zhang2022provably}, for the S-shaped~\cite{henriksen2020efficient}, $Sigmoid(x)\cdot Tanh(y),x\cdot Sigmoid(y)$~\cite{prover,ko2019popqorn}, and ReLU functions~\cite{singh2019abstract,boopathy2019cnn}. 
These neuron-wise tightest linear bounds can empirically give rise to certified robustness results and stand for the highest precision among other relevant work~\cite{meng2022adversarial}. 
However, they focus on the ReLU and Sigmoid layer and ignore multi-variate functions like MaxPool, which is widely adopted in CNNs~\cite{liu2022convnet,xie2017aggregated} yet is far more complex to verify.

Until recently, some works attempt to make non-trivial linear relaxation for MaxPool~\cite{boopathy2019cnn,singh2019abstract,lorenz2021robustness}  or transform  MaxPool into a sequence of affine and ReLU layers to verify~\cite{osip}. 
Unfortunately, they both certify robustness results coarsely and some consume much time to gain non-optimal results~\cite{osip,lorenz2021robustness}. These works fail to capture the neuron-wise tightest linear bounds for MaxPool, leaving ample room for the linear bounds to be further adapted to tighten. 
Recently, a block-wise tightest linear upper bound was proposed for MaxPool~\cite{maxlin2024}, which minimizes the over-approximation for the ReLU+MaxPool block. However, this upper bound is the block-wise tightest only when the verifier uses a single upper linear constraint to bound the ReLU. 
When ReLU is bounded by BaB techniques and has no precision loss, the neuron-wise linear bounds achieve higher precision than the block-wise tightest linear upper bound~\cite{maxlin2024} for the ReLU+MaxPool block.

\begin{figure*}
\centering
\includegraphics[width=1.85\columnwidth]{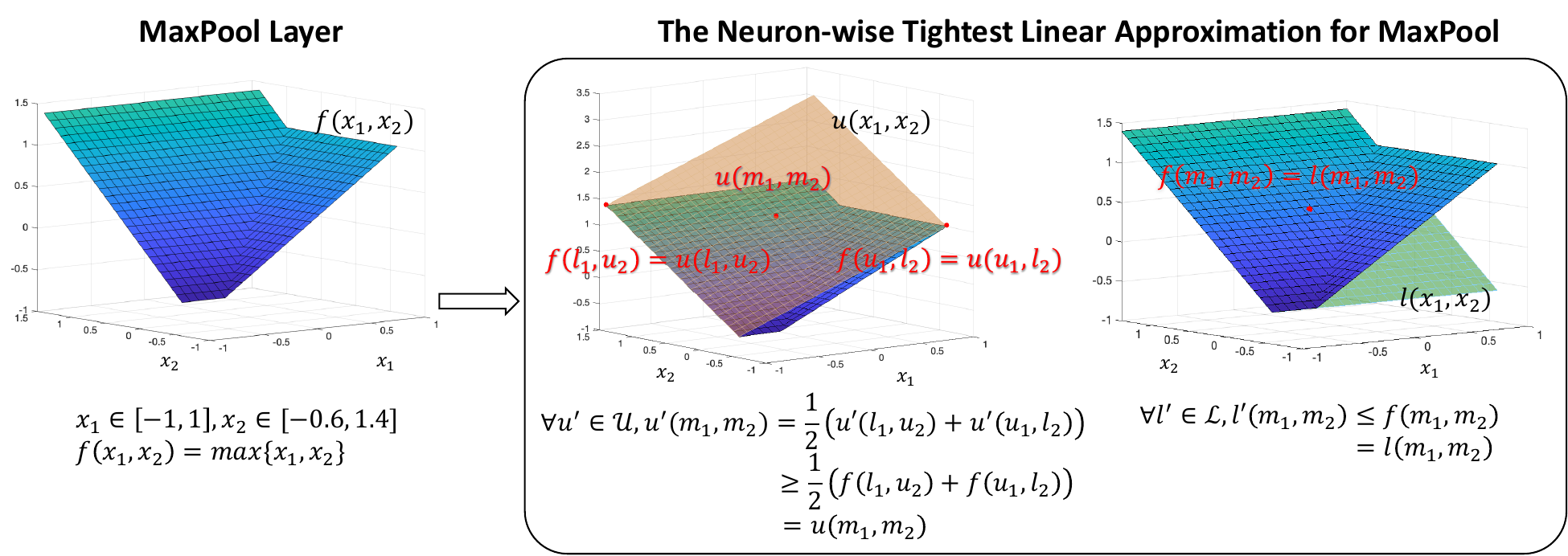} 
\caption{An illustration for the neuron-wise tightest linear bounds for bivariate MaxPool function. From left to right, the subfigures are the MaxPool function, upper linear bound plane $u(x_1,x_2)$, and lower linear bound plane $l(x_1,x_2)$, respectively. The red dots are $u(l_1,u_2), u(u_1,l_2), u(m_1,m_2),$and $l(m_1,m_2)$, where $m_i=\frac{1}{2}(l_i+u_i), i=1,2$. $\mathcal{L}$ and $\mathcal{U}$ are the sets of all lower and upper linear bounds for MaxPool, respectively. }
\label{fig-intro}
\end{figure*}
To address the above challenges, we propose Ti-Lin, a robustness verifier for MaxPool-based CNNs with tight linear approximation. To the best of our knowledge, we are the first to propose the neuron-wise tightest linear bounds for MaxPool.
Specifically, we introduce the notion of neuron-wise tightest and utilize the characteristic of piece-wise linearity of MaxPool $f(\boldsymbol{x})$.
We make the upper linear bound $u(\boldsymbol{x})$ passing two points $f(\boldsymbol{a})$ and $f(\boldsymbol{b})$, where $\boldsymbol{a}$ and $\boldsymbol{b}$ are the endpoints of the space diagonal for MaxPool's input region, and the lower linear bound passing $f(\boldsymbol{m})$.
A simple example are shown in Figure~\ref{fig-intro}, where $\boldsymbol{a}=(l_1,u_2)$ and $\boldsymbol{b}=(u_1,l_2)$. 
We prove that our proposed linear approximation is the neuron-wise tightest and its computation is efficient with the time complexity $\mathcal{O}(pool\_size\times log3)$, where $pool\_size$ is the size of inputs to be pooled. Moreover, Ti-Lin can be easily integrated into different state-of-the-art verifiers, such as ERAN~\cite{Convex_Relaxation_Barrier,singh2018fast,singh2019boosting,muller2022prima,singh2019abstract,lorenz2021robustness} and $\alpha,\beta$-CROWN~\cite{xu2020fast,wang2021beta,tjeng2018evaluating,pmlr-v162-zhang22ae,Zhang2022GeneralCP,zhang2018efficient}, the VNN-COMP 2021-2023 winner~\cite{bak2021second,muller2022third,brix2023fourth}. The integration allows Ti-Lin to certify different types of MaxPool-based networks (e.g., CNNs and PointNets) against $l_1,l_2,l_\infty$-norm perturbations.

To illustrate the superiority of Ti-Lin, we integrate Ti-Lin into four verification frameworks, including CNN-Cert~\cite{boopathy2019cnn}, DeepPoly~\cite{singh2019abstract}, ERAN,  and $\alpha,\beta$-CROWN, and evaluate Ti-Lin 
with open-sourced benchmarks on the MNIST~\cite{lecun1998mnist}, CIFAR-10~\cite{krizhevsky2009learning}, Tiny ImageNet~\cite{krizhevsky2009learning}, and ModelNet40~\cite{wu20153d} datasets.
The experiment results show that Ti-Lin outperforms the state-of-the-art verifiers for MaxPool-based networks, including CNN-Cert~\cite{boopathy2019cnn}, DeepPoly~\cite{singh2019abstract}, and 3DCertify~\cite{lorenz2021robustness}, MN-BaB, ERAN, and $\alpha,\beta$-CROWN, MaxLin~\cite{maxlin2024}, OSIP~\cite{osip}, with up to 69.0\%, 39.4\%, 7.1\%, 39.4\%, 7.1\%, 78.6\%, 28.6\% and 65.0\% improvement in terms of tightness, respectively. Meanwhile, having almost the same time cost as the fastest tool, Ti-Lin incurs almost no additional time cost and has up to 5.3$\times$, 1.7$\times$,  and 420.9$\times$ speedup over 3DCertify, ERAN, and OSIP, respectively.

In summary, our work proposes a neuron-wise tightest linear approximation for MaxPool, Ti-Lin, with high time efficiency, which works for various CNNs and different verification frameworks. 
By tightening linear approximation for MaxPool, our approach outperforms the state-of-the-art tools with up to 78.6\% improvement to the certified results and has the same time efficiency as the fastest baseline method.

\section{Related work}
\label{section2}
\paragraph{Adversarial attacks and defenses}
Many research studies have shown that machine learning models are vulnerable to adversarial examples~\cite{chen2017zoo,chen2020hopskipjumpattack,carlini2017towards,silva2020opportunities,wu2023adversarial,DBLP:conf/nips/0008XZZMC0Z23,DBLP:conf/nips/ZhangRTW23,DBLP:conf/nips/JiaYSNRRLP23}, which pose severe concerns for their deployment in security and safety-critical applications such as autonomous driving. With the existence of adversarial attacks, a plethora of research on adversarial defense has been created~\cite{goodfellow2014explaining,madry2018towards,katz2017reluplex,katz2019marabou,cohen2019certified,henriksen2020efficient,wu2021tightening,xu2020fast,wang2021beta,lyu2020fastened,balunovic2019certifying}. However, empirical defenses~\cite{papernot2016distillation,buckman2018thermometer} cannot provide a formal robustness guarantee and are often compromised by adaptive, unforeseen attacks~\cite{carlini2017adversarial,athalye2018obfuscated,ghiasi2020breaking}. Therefore, our work focuses on certified defense which can provide a provably safe guarantee, and targets MaxPool-based convolutional neural networks, which are widely used for image classification.

\paragraph{Robustness verification for MaxPool-based neural networks}
Recently, Some work based on mixed integer linear programming  and satisfiability modulo theory~\cite{katz2019marabou,mixed-integer_programming,Piece-Wise_Linear} proposed a complete verification framework that supports CNNs with arbitrary piece-wise linear activation functions. However, it cannot apply to MaxPool-based networks with other activation functions, such as Sigmoid. Recently, PRIMA~\cite{muller2022prima} proposes a novel verifier based on multi-neuron relaxation for arbitrary, bounded, multivariate non-linear activation functions. MN-BaB~\cite{ferrari2022complete} propose a state-of-the-art complete neural network verifier that builds on the tight multi-neuron constraints proposed in PRIMA.
Another line of research based on linear approximation has the advantage of high time efficiency and, meanwhile, preserves high precision of certifying the robustness of CNNs. 
DeepPoly~\cite{singh2019abstract} gives non-trivial linear bounds to MaxPool and its parallel work, CNN-Cert~\cite{boopathy2019cnn} leverages a hyperplane containing $n$ different particular points to give the upper and lower bound. 3DCertify~\cite{lorenz2021robustness} gives tighter bounds than DeepPoly by the double description method~\cite{fukuda2005double}. Recently, OSIP~\cite{osip} verifies MaxPool-based networks by transforming the MaxPool layer into a sequence of affine and ReLU layers. However, the above techniques verify the robustness results coarsely and some methods waste much time to gain sub-optimal results. Recently, MaxLin~\cite{maxlin2024} proposed provably block-wise tightest upper linear constraints for ReLU+MaxPool blocks. However, this block-wise tightest property holds only when the ReLU is bounded by a single upper linear constraint, rather than by much tighter constraints, such as those provided by BaB techniques.

\section{Preliminaries}
\label{section3}
\subsection{Certified robustness bound}
Let $F(\boldsymbol{x}):\mathbb{R}^{n_0}\to\mathbb{R}^{n_K}$ be a neural network classifier function with (K+1) layers and $\boldsymbol{x_0}$ be an input data point. $t$ denotes the true label of $\boldsymbol{x_0}$ in $F$. Let $\mathbb{B}_p(\boldsymbol{x_0},\epsilon)$ denotes $\boldsymbol{x_0}$ perturbed within an $l_p$-norm ball with radius $\epsilon$, that is $\mathbb{B}_p(\boldsymbol{x_0},\epsilon)=\{\boldsymbol{x}|\Vert \boldsymbol{x}-\boldsymbol{x_0}\Vert_p \leq \epsilon\}$.  In this work, $p=1,2,\infty$. Next, we introduce the notion of local robustness bound $\epsilon_r$, which can be certified by complete verifiers. 
\begin{definition}[Local robustness bound]
\label{local}
If $\epsilon_r (\epsilon_r\geq0)$ is the local robustness bound of an input $\boldsymbol{x_0}$ in the neural network, if and only if these two conditions (i) $argmax_i F_i(\boldsymbol{x})=t,\forall \boldsymbol{x}\in \mathbb{B}_p(\boldsymbol{x_0},\epsilon_r)$
 and (ii) $\forall \delta>0,\exists \boldsymbol{x_a}\in \mathbb{B}_p(\boldsymbol{x_0},\epsilon+\delta)s.t. \mathop{\arg\max}\limits_{i}F_i(\boldsymbol{x_a})\neq t$ are satisfied.
\end{definition}

Apparently, $\epsilon_r$ is the maximum absolute safe radius of $\boldsymbol{x_0}$. Although computing $\epsilon_r$ is an essential problem, it is computationally expensive and  NP-complete
~\cite{katz2017reluplex}. Thus, it is practical to compute a bound that satisfies condition (ii) of Definition~\ref{local} but is lower than $\epsilon_r$.

\begin{definition}[Certified lower bound]
If $\epsilon_{cert}(\epsilon_{cert}\geq0)$ is a certified lower bound of input $\boldsymbol{x_0}$  in a neural network,  if and only if these two conditions (i) $\epsilon_{cert}\leq \epsilon_r$ and (ii) $\mathop{\arg\max}\limits_{i} F_i(\boldsymbol{x})=t,\forall  \boldsymbol{x}\in \mathbb{B}_p(\boldsymbol{x_0},\epsilon_r)$ are satisfied.
\end{definition}
Robustness lower bound can be certified by incomplete verifiers. As incomplete verifiers risk precision loss to efficiently certify the robustness of more types of networks, the value of $\epsilon_{cert}$ is a key criterion to evaluate the tightness of robustness verification methods.

\subsection{Linear approximation}
Define $\boldsymbol{l}, \boldsymbol{u}$ are the lower and upper bounds of the input of the intermediate layer, whose function is $f(\cdot)$; that is,  $\boldsymbol{x}\in[\boldsymbol{l},\boldsymbol{u}]$. 
The essence of the linear approximation technique is giving linear bounds to every layer, that is 
\begin{equation}\label{eq-sound}
    \forall \boldsymbol{x}\in [\boldsymbol{l},\boldsymbol{u}], l(\boldsymbol{x})\leq f(\boldsymbol{x})\leq u(\boldsymbol{x})
\end{equation}
\begin{definition}[Upper/Lower linear bound]
Let $f_i(\boldsymbol{x})$ be the function of the $i$-th neuron in the intermediate layer $f(\boldsymbol{x})$ of neural network $F$, with $\boldsymbol{x}\in[\boldsymbol{l},\boldsymbol{u}]\subset \mathbb{R}^{n},\boldsymbol{a_u},\boldsymbol{a_l}\in\mathbb{R}^{n}, b_u,b_l\in\mathbb{R}$ and $$u_i(\boldsymbol{x})=\boldsymbol{a_u}\boldsymbol{x}+b_u,l_i(\boldsymbol{x})=\boldsymbol{a_l}\boldsymbol{x}+b_l$$ $u_i(\boldsymbol{x})$ and $l_i(\boldsymbol{x})$ are linear upper and lower bounds of $f_i(x)$ if $l_i(\boldsymbol{x})\leq f_i(\boldsymbol{x})\leq u_i(\boldsymbol{x}),\forall \boldsymbol{x}\in[\boldsymbol{l},\boldsymbol{u}]$. 
\end{definition} 

Here, $\boldsymbol{a_u}$ and $ b_u$ are the slope and intercept of the upper linear bound $u_i(\boldsymbol{x})$, respectively.  $n$ equals to the input size of $f_i(\boldsymbol{x})$, e.g., when $f_i(\boldsymbol{x})$ is MaxPool and the pool size is $2\times2$, $n$ equals to 4.

\subsection{Computing certified results}
Robustness is typically measured by certified accuracy and certified robustness bounds.

\subsubsection{Certified accuracy}
Certified accuracy is widely used in most verifiers~\cite{osip,Zhang2022GeneralCP,singh2019abstract,muller2022prima} and represents the percentage of the instances verified as safe.
In essence, the robustness verifier of neural networks aims to compute the bounds of the output neurons within a perturbed input region.
That is, after verification, we would get $$\boldsymbol{L} \leq F(\boldsymbol{x})\leq \boldsymbol{U},\forall x\in \mathbb{B}_p(\boldsymbol{x_0},\epsilon)$$
where  $\boldsymbol{L}$ and $\boldsymbol{U}$ are the lower and upper bounds of the whole network $F(\boldsymbol{x})$, respectively.  As $F(\boldsymbol{x}): \mathbb{R}^{n_0}\to\mathbb{R}^{n_K}$, we assume $\boldsymbol{L}=(L_1,L_2,\cdots,L_{n_K})$ and $\boldsymbol{U}=(U_1,U_2,\cdots,U_{n_K})$. $[n]$ denotes $\{1,2,\cdots,n\}$.
    Then, if 
    \begin{equation}\label{eq-globallowerbound}
        L_t>U_i, \forall i\neq t, i \in [n_K]
    \end{equation}
    we can conclude that $\epsilon$ is a safe perturbation radius, i.e., $\boldsymbol{x_0}$ is verified as safe under the perturbation whose radius is  $\epsilon$ and form is $l_p$ norm. Otherwise, $\boldsymbol{x_0}$ is verified as unknown. The certified accuracy is computed as the proportion of inputs verified as safe relative to the total number of inputs. 

\subsubsection{Certified robustness bounds}\label{section-compute}

Some verification methods~\cite{boopathy2019cnn,wu2021tightening} would use the binary search algorithm to compute the maximal certified lower bounds for input $\boldsymbol{x_0}$, a.k.a. the certified robustness bounds, as their robustness results. 
Concretely,  the perturbation range is initialized as $\epsilon$. Given an input $\boldsymbol{x_0}$ and neural network $F$, the verification framework computes the bounds for the output layers within the input region $\boldsymbol{x}\in\mathbb{B}_p(\boldsymbol{x_0},\epsilon)$. $\boldsymbol{L}$ and $\boldsymbol{U}$ represent the lower and upper bounds of neural network $F$, respectively. 
When the perturbation range is certified safe, the perturbation ranges $\epsilon$ would be increased. When $\epsilon$ cannot be certified safe,   $\epsilon$ would be decreased.  
Finally, when the above verification process is repeated, the search range is gradually reduced. 
After repeated finite times,  the algorithm terminates and returns $\epsilon$ as the maximal certified lower bound that the verifier can verify for the input $\boldsymbol{x_0}$.

\section{Ti-Lin: A robustness verifier for MaxPool-based networks}
\label{section4}
In this section, we present Ti-Lin, the neuron-wise tightest and easy-to-integrate linear approximation for MaxPool.

\subsection{Tight linear approximation for MaxPool}
 We assume that $n$ is the size of inputs that will be pooled. Then, we use $f(x_1,\cdots,x_n)=max\{x_1,\cdots,x_n\}, x_i\in[l_i,u_i]$ to represent the MaxPool function without loss of generality. 

\begin{theorem}
\label{theoremMaxPool}
Given $f(x_1,\cdots,x_n)=max\{x_1,\cdots,x_n\}, x_i\in[l_i,u_i]$, we first select the first, second, and third largest elements of $\{u_i|i=1,\cdots,n\}$, whose indexes are denoted as $i,j,k$, respectively. We then choose the largest element of $\{l_i|i=1,\cdots,n\}$ and denote it as $l_{max}$. Define $\boldsymbol{m}=(m_1,\cdots,m_n)=(\frac{l_1+u_1}{2},\cdots,\frac{l_n+u_n}{2})\in\mathbb{R}^n$. Then the linear bounds of the MaxPool function are:

\textbf{Upper linear bound:} 

$u(x_1,\cdots,x_n)=\sum_ia_i(x_i-l_i)+b$. Specifically, there are four different cases:

Case 1:
if $(l_i=l_{max}) \wedge (l_i\geq u_j)$, then $a_i=1;b=l_i;a_q=0,q\neq i$.

Case 2:
if $(l_i=l_{max}) \wedge (u_j > l_i \geq u_k) $, then $a_i=1;a_j=\frac{u_j-l_i}{u_j-l_j};b=l_i;a_q=0,q\neq i,j$.

Case 3:
if $(l_j=l_{max})\wedge (l_i\neq l_{max})\wedge (u_j \geq l_j \geq u_k)$,  then $a_i=\frac{u_i-l_j}{u_i-l_i};a_j=1;b=l_j;a_q=0,q\neq i,j$.

Case 4:
otherwise,  $a_i=\frac{u_i-u_k}{u_i-l_i};a_j=\frac{u_j-u_k}{u_j-l_j};b=u_k;a_q=0,q\neq i,j$.

\textbf{Lower linear bound:}

$l(x_1,\cdots,x_n)=x_q,q=\mathop{\arg\max}\limits_{i} (m_i)$
\end{theorem}
We prove that the linear bounds in Theorem~\ref{theoremMaxPool} are sound in the Appendix, that is, $l(\boldsymbol{x})\leq f(\boldsymbol{x})\leq u(\boldsymbol{x}), \forall \boldsymbol{x}\in[\boldsymbol{l},\boldsymbol{u}]$.

\subsection{Neuron-wise tightest property}
\begin{figure*}[!t]
\centering
\includegraphics[width=1.3\columnwidth]{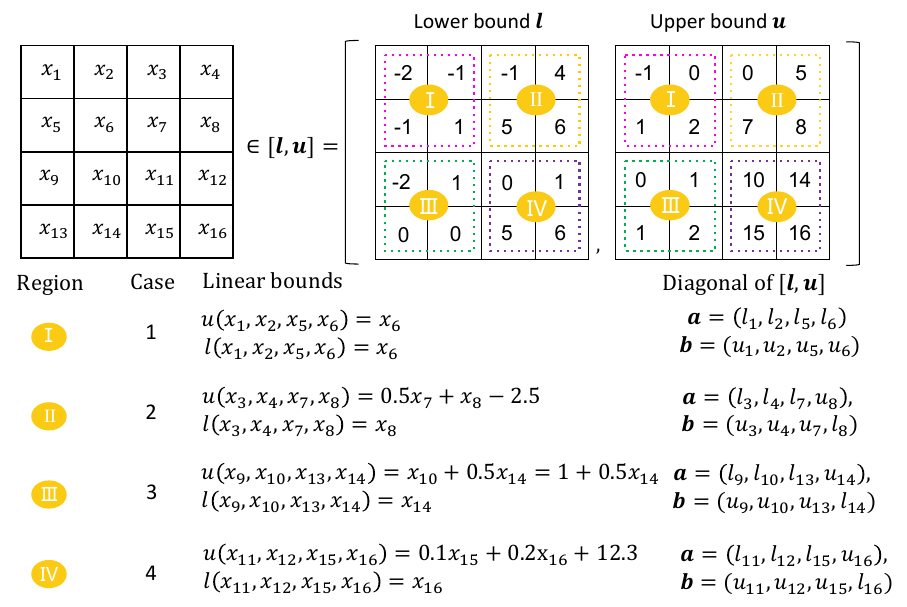} 
\caption{Toy examples of the neuron-wise tightest linear bounds for the four cases in Theorem~\ref{theoremMaxPool}.}
\label{fig-example}
\end{figure*}
Linear approximation technique inevitably introduces an over-approximation zone of the intermediate layer's output value to keep sound, according to Equation~\ref{eq-sound}. 
As for single non-linear layer positioned between linear layers, the over-approximation zone between the upper and lower linear constraints will be propagated throughout the verification process, and affect the whole precision loss. 
Therefore, it is necessary to minimize the over-approximation zone for every non-linear layer, such as, MaxPool.
Furthermore, for cases where two non-linear layers, such as an adjacent ReLU and MaxPool block, are positioned between linear layers, it is also more effective to minimize the over-approximation zone for MaxPool when the activation is bounded by the precise constraints provided by BaB techniques.

We now introduce the concept of neuron-wise tightest~\cite{zhang2022provably}, which denotes the characteristic of linear bounds that yield the most compact over-approximation zone for the non-linear neuron.
\begin{definition}[Neuron-wise Tightest]
\label{d1}
Upper linear bound $u:[\boldsymbol{l},\boldsymbol{u}]$ $\to \mathbb{R}$ and lower linear bound $l:[\boldsymbol{l},\boldsymbol{u}]$ $\to \mathbb{R}$ are the neuron-wise tightest if and only if both  
$\iint_{[\boldsymbol{l},\boldsymbol{u}]}(u(\boldsymbol{x})-l(\boldsymbol{x}))d\boldsymbol{x}$ are the minimal.
\end{definition}

The upper and lower bounds of the intermediate layer's input value rely on the verification process of the previous layers and decide the input region of the current layer's linear approximation. 
Based on this, we can get the general form of the neuron-wise tightest linear bounds for all non-linear, bounded, and continuous functions in neural networks.

 \begin{theorem}
\label{theorem-generalbound}
Given $f$ is continuous and bounded, we define $\boldsymbol{m}=\frac{\boldsymbol{l}+\boldsymbol{u}}{2}$. Then, (i) if $u(\boldsymbol{m})$  reaches the minimum, then the upper linear bound for $f$ is the neuron-wise tightest; (ii) if $l(\boldsymbol{m})$ reaches the maximum, then the lower linear bound for $f$ is the neuron-wise tightest.
\end{theorem}
The proof of Theorem~\ref{theorem-generalbound}  is put in the Appendix due to page limit.

Existing methods~\cite{lorenz2021robustness,singh2019abstract,boopathy2019cnn,osip} ignore the piece-wise linear property of MaxPool and fail to capture the tightest linear approximation for MaxPool. Their resulting coarse linear approximation leads to non-optimal robustness results. In this paper, we can prove that our proposed linear approximation in Theorem~\ref{theoremMaxPool} is the neuron-wise tightest and produces the smallest over-approximation zone.
\begin{theorem}
\label{tightest}
    The MaxPool linear bounds from Theorem~\ref{theoremMaxPool} are the neuron-wise tightest.
\end{theorem}

The proof of Theorem~\ref{tightest}  is put in the Appendix due to page limit.

To explain the provably neuron-wise tightness clearly, we provide a toy example for  Theorem~\ref{theoremMaxPool} in Figure~\ref{fig-example}.
In this example, we consider a MaxPool layer with a pool size of $2\times2$, a padding size of P=[0 0 0 0], and a stride S=[2 2] to clarify the use of Ti-Lin. The MaxPool operation is applied on four regions: I, II, III, and IV, and we give an example for each case in Theorem~\ref{theoremMaxPool}, as shown in Figure~\ref{fig-example}. 
According to the values of $\boldsymbol{l}$ and $\boldsymbol{u}$, we can give the linear bounds for each region as shown in Figure~\ref{fig-example}. We list all $\boldsymbol{a}$ and $\boldsymbol{b}$ for each case, which are the endpoints of the space diagonal of the input region and can be formally defined as:
$$\boldsymbol{a}:=\mathbb{I}\cdot \boldsymbol{l}+(\boldsymbol{1}-\mathbb{I})\cdot \boldsymbol{u}$$
$$\boldsymbol{b}:=\boldsymbol{l}+\boldsymbol{u}-\boldsymbol{a}=(\boldsymbol{1}-\mathbb{I})\cdot \boldsymbol{l}+\mathbb{I}\cdot \boldsymbol{u}$$
where $\cdot$ is dot product and $\mathbb{I}$ represents a indicator matrix that has the same shape as $\boldsymbol{l}$. From Figure~\ref{fig-example}, it is easy to get $f(\boldsymbol{a})=u(\boldsymbol{a}), f(\boldsymbol{b})=u(\boldsymbol{b})$. Then, leveraging the linearity of the upper linear bounds $u(\boldsymbol{x})$, we have $ \forall u'\in\mathcal{U}, u(\boldsymbol{m})=\frac{1}{2}(u(\boldsymbol{a})+u(\boldsymbol{b}))
     =\frac{1}{2}(f(\boldsymbol{a})+f(\boldsymbol{b}))
     \leq\frac{1}{2}(u'(\boldsymbol{a})+u'(\boldsymbol{b}))
     =u'(\frac{1}{2}(\boldsymbol{a}+\boldsymbol{b}))
     =u'(\boldsymbol{m})$. Then, we prove that the upper linear bound is neuron-wise tightest, according to  Theorem~\ref{theorem-generalbound}.
     In terms of lower linear bound, we have
     $ \forall l'\in\mathcal{L}, l(\boldsymbol{m})=f(\boldsymbol{m})
      \geq l'(\boldsymbol{m})$.
Then, we can prove that the lower linear bound is neuron-wise tightest, according to  Theorem~\ref{theorem-generalbound}.

The time complexity of Theorem~\ref{theoremMaxPool} depends on selecting the three largest upper bounds. The time complexity of computing the linear bound for one region is $\mathcal{O}(pool\_size\times log3)$, where $pool\_size$ is the size of inputs to be pooled. For example, in Figure~\ref{fig-example}, the MaxPool layer operates on a kernel of $2 \times 2$ and thus, $pool\_size=2\times 2=4$.

  \section{Experimental evaluation}
  \label{section5}

\begin{table*}[htbp]
\centering
\caption{The performance of Ti-Lin on DeepPoly verifier (backsubstitution-based). }

\huge
\resizebox{0.80\textwidth}{!}{
\begin{threeparttable}
\begin{tabular}{clrrrrrrrr}
\toprule
  & &  \multicolumn{3}{c}{{Certified Accuracy (\%)}}&UB (\%)$^{\ddag}$&\multicolumn{3}{c}{{Average Runtime(s)} }&\multicolumn{1}{c}{{Speedup} }
    \\\cmidrule(lr){6-6}\cmidrule(lr){7-9}\cmidrule(lr){3-5}\cmidrule(lr){10-10}
    
    \multicolumn{1}{c}{{Dataset}}&  \multicolumn{1}{c}{{Network }}&  \multirow{1}{*}{{ DeepPoly }} &  \multirow{1}{*}{{ 3DCertify$^{\dagger}$ }} & \multicolumn{1}{c}{{Ti-Lin$^{\dagger}$}}&&\multirow{1}{*}{{DeepPoly}}  & \multirow{1}{*}{{3DCertify}}& \multicolumn{1}{c}{{Ti-Lin}}   &\multirow{1}{*}{{vs.  3DCertify  } }  \\
  \midrule
  \multirow{7}{*}{\shortstack{MNIST}}
  &Conv\_MaxPool&39.5&50.6&\textbf{53.1}&93.8&0.9&2.6&1.4&\textbf{1.9}\\
  &Convnet\_MaxPool&3.1&3.1&\textbf{4.2}&71.9&0.2&0.3&0.2&\textbf{1.5}\\
  &CNN, 4 layers&79.8&86.2&\textbf{89.4}&98.9&2.5&6.7&4.6&\textbf{1.5}\\
  &CNN, 5 layers&90.9&90.9&\textbf{92.9}&100.0&13.5&23.2&13.8&\textbf{1.7}\\
  &CNN, 6 layers&78.8&\textbf{85.9}&\textbf{85.9}&99.0&23.9&56.0&27.0&\textbf{2.1}\\
  &CNN, 7 layers&81.3&86.8&\textbf{90.1}&100.0&32.7&71.1&44.7&\textbf{1.6}\\
  &CNN, 8 layers&38.4&70.7&\textbf{77.8}&99.0&30.1&76.8&41.7&\textbf{1.8}\\
\midrule
  \multirow{6}{*}{\shortstack{CIFAR-10}}
  &Conv\_MaxPool&0.0&0.0&\textbf{2.1}&93.9&19.7&30.3&17.8&\textbf{1.7}\\
  &CNN, 4 layers&59.3&66.7&\textbf{70.4}&100.0&8.6&14.2&10.9&\textbf{1.3}\\
  &CNN, 5 layers&68.8&68.8&\textbf{75.0}&87.5&36.2&55.5&35.4&\textbf{1.6}\\
  &CNN, 6 layers&\textbf{85.7}&\textbf{85.7}&\textbf{85.7}&100.0&58.2&115.5&55.4&\textbf{2.1}\\
  &CNN, 7 layers&13.8&34.5&\textbf{37.9}&72.7&115.1&107.7&95.6&\textbf{1.1}\\
  &CNN, 8 layers&41.7&\textbf{50.0}&\textbf{50.0}&91.7&60.5&128.9&55.8&\textbf{2.3}\\
  \midrule
\multirow{5}{*}{\shortstack{ModelNet40}}
&16p\_Natural & 74.6 & 74.6 & \textbf{76.3} & \multicolumn{1}{r}{  -}&7.3 & 13.0 & 7.7 & \textbf{1.7} \\
&32p\_Natural & 56.5 & 56.5 & \textbf{58.1} &\multicolumn{1}{r}{  -}& 13.2& 25.4 & 14.0 & \textbf{1.8}\\
&64p\_Natural & 28.1 & 28.1 & \textbf{29.7} &\multicolumn{1}{r}{-}& 32.1& 81.2 & 30.8 & \textbf{2.6}\\
&64p\_FGSM&12.1&12.1&\textbf{13.6}&\multicolumn{1}{r}{-}&27.8&177.0&42.3&\textbf{4.2}\\
&64p\_IBP&\textbf{88.3}&\textbf{88.3}&\textbf{88.3}&\multicolumn{1}{r}{-}&29.3&174.9&32.8&\textbf{5.3}\\

\bottomrule
\end{tabular}
    \begin{tablenotes}
        \item $^{\dagger}$ 
3DCertify is built atop the DeepPoly framework, which is why we integrate Ti-Lin into DeepPoly for comparison. This integration allows us to evaluate their performance within the same framework.
        \item $^{\ddag}$ UB represents the upper bounds for robustness verification computed by PGD attacks~\cite{PGDattack}. As PGD attacks is not applicable to PointNets, the UB results of PointNets is - .
    \end{tablenotes}
    \end{threeparttable}
}
\label{table-3dcertify}
\end{table*}

We conduct extensive experiments on CNNs by comparing Ti-Lin to seven SOTA verifiers, including CNN-Cert~\cite{boopathy2019cnn}, DeepPoly~\cite{singh2019abstract}, 3DCertify~\cite{lorenz2021robustness}, MN-BaB~\cite{ferrari2022complete}, ERAN~\cite{singh2018fast,singh2019boosting,muller2022prima,singh2019abstract,lorenz2021robustness}, OSIP~\cite{osip}, $\alpha,\beta$-CROWN~\cite{xu2020fast,wang2021beta,tjeng2018evaluating,pmlr-v162-zhang22ae,Zhang2022GeneralCP,zhang2018efficient}, the VNN-COMP 2021-2023 winner~\cite{bak2021second,muller2022third,brix2023fourth}, and MaxLin~\cite{maxlin2024}.
All experiments were conducted on a server with a 48core Intel Xeon Silver 4310 CPU and 125 GB RAM.
  
  \subsection{Experimental setup}
\begin{table*}[htbp]
\centering
\caption{The performance of Ti-Lin on ERAN verifier (MN-based). }
  \resizebox{0.80\textwidth}{!}{
  \begin{threeparttable}
  \begin{tabular}{clrrrrrrrr}
\toprule
 & &  \multicolumn{3}{c}{{Certified Accuracy (\%)}}&UB (\%)&\multicolumn{3}{c}{{Average Runtime(min)} }& \multicolumn{1}{c}{Speedup} 
    \\\cmidrule(lr){7-9}\cmidrule(lr){3-5}\cmidrule(lr){10-10}\cmidrule(lr){6-6}

          \multicolumn{1}{c}{{Dataset}}&  \multicolumn{1}{c}{{Network }} &  \multirow{1}{*}{{ MN-BaB }}     &\multirow{1}{*}{{ ERAN }} & \multicolumn{1}{c}{{Ti-Lin$^{\dagger}$}}&&   \multirow{1}{*}{{MN-BaB}}  &\multirow{1}{*}{{ERAN}}& \multicolumn{1}{c}{{Ti-Lin}}   & \multicolumn{1}{c}{vs. ERAN}    \\
\midrule
  \multirow{8}{*}{\shortstack{MNIST}}
  &Conv\_MaxPool&38.3&56.8&\textbf{58.0}&93.8&1.8 & 1.2 & 0.7 & \textbf{1.7}\\
  &ConvNet\_MaxPool&3.1&\textbf{6.3}&\textbf{6.3}&71.9&2.5 & 0.6 & 0.4 & \textbf{1.6}\\
  &CNN, 4 layers&79.8&87.2&\textbf{89.4}&98.9&1.8 & 0.4 & 0.3 & \textbf{1.4}\\
  &CNN, 5 layers&90.9&91.9&\textbf{92.9}&100.0&1.5 & 0.8 & 0.5 & \textbf{1.4}\\
  &CNN, 6 layers&78.8&\textbf{85.9}&\textbf{85.9}&99.0&1.9 & 2.3 & 2.1 & \textbf{1.1}\\
  &CNN, 7 layers&81.3&86.8&\textbf{90.1}&100.0&1.6 & 3.0 & 1.9 & \textbf{1.5}\\
  &CNN, 8 layers&38.4&70.7&\textbf{77.8}&99.0&5.0 & 7.4 & 6.1 & \textbf{1.2}\\
\midrule
  \multirow{6}{*}{\shortstack{CIFAR-10}}
  &Conv\_MaxPool&0.0&0.0&\textbf{2.1}&93.9&5.0  & 17.3 & 13.1 & \textbf{1.3}\\
  &CNN, 4 layers&59.3&66.7&\textbf{70.4}&100.0&3.7 & 5.8  & 5.6  & \textbf{1.0}\\
  &CNN, 5 layers&68.8&68.8&\textbf{75.0}&87.5&2.9  & 2.9  & 2.0  & \textbf{1.4}\\
  &CNN, 6 layers&\textbf{85.7}&\textbf{85.7}&\textbf{85.7}&100.0&2.9  & 5.1  & 3.1  & \textbf{1.6}\\
  &CNN, 7 layers&13.8&34.5&\textbf{37.9}&72.7&6.8  & 9.0  & 7.1  & \textbf{1.3}\\
  &CNN, 8 layers&41.7&50.0&\textbf{58.3}&91.7&4.9  & 11.1 & 9.2  & \textbf{1.2}\\
\bottomrule
\end{tabular}
    \begin{tablenotes}
        \item $^{\dagger}$ MN-BaB uses PRIMA's BaB Bound for MaxPool and employs a dual solver in the BaB solver, which makes it difficult to integrate Ti-Lin directly. However, both MN-BaB and ERAN use MN bounding for ReLU. Therefore, in this table, we integrate Ti-Lin into ERAN for comparison.
    \end{tablenotes}
    \end{threeparttable}
}
\label{table-eran}
\end{table*}

\noindent
   \textbf{Framework. }
Our proposed neuron-wise tightest linear bounds for MaxPool are independent of the concrete verifier. For a fair comparison, we compare Ti-Lin to the baseline methods on the same frameworks. Concretely, we instantiate Ti-Lin on three types of verifiers, including
the backsubstitution-based verifiers (CNN-Cert, DeepPoly, and 3DCertify), the multi-neuron abstraction (MN) based verifier (ERAN), and the Branch and Bound (BaB) based verifier ($\alpha,\beta$-CROWN). Concretely, backsubstitution~\cite{singh2019abstract}, a.k.a. bound propagation,  uses linear approximation to compute the bounds of the output neurons and achieves a good balance between precision and efficiency. MN~\cite{singh2019beyond,muller2022prima} leverages the interdependencies between neurons and computes optimal convex relaxations for the output of k-ReLU operations jointly for improved precision. BaB~\cite{Zhang2022GeneralCP}  computes the sound linear bounds for the non-linear layers and recursively splits the layer's input region with added constraints to achieve increasingly tighter bounds.

\noindent
   \textbf{Benchmarks and experimental settings. }
  We follow the settings used in the baseline methods.
 MNIST~\cite{lecun1998mnist}, CIFAR-10~\cite{krizhevsky2009learning}, Tiny ImageNet~\cite{deng2009imagenet}, and ModelNet40~\cite{wu20153d} are used in experiments. 
All networks in experiments are open-sourced and come from 3DCertify, ERAN, VNN-COMP2021~\cite{bak2021second}, and CNN-Cert. We follow the timeout setting in VNN-COMP2021 and $\alpha,\beta$-CROWN, timeout is  180  seconds for all experiments. Other detailed experiment settings are in the Appendix.

\noindent
   \textbf{Metric. }
    We follow the metrics used in the baseline methods.
In terms of effectiveness, we use the certified accuracy as our metric. 
In terms of efficiency, we record the average runtime over all instances. We use $t'/t$ to represent the relative time improvement of Ti-Lin over the baseline methods, where $t$ and $t'$ denote the average computation time of Ti-Lin and the baseline methods. On the $\alpha,\beta$-CROWN framework, we also record the timeout rate. \textit{\textbf{It is worth noting that, we only compute the speedup of Ti-Lin over 3DCertify, ERAN, and OSIP}}, as both  3DCertify and ERAN MaxPool linear bounds rely on the double description method~\cite{fukuda2005double} to search for optimal linear bounds, which has high time complexity for computing its MaxPool linear approximation compared to other methods. While
OSIP bounds MaxPool by transforming it into a sequence of ReLU and affine layers. However, the additional layers introduced by this transformation significantly increase the computational time.
As for  other methods, whose linear approximation for MaxPool is directly obtained by sorting the upper and lower bounds, operate similarly to our proposed method, and the speedup of Ti-Lin is primarily due to its high precision rather than its low time complexity. Therefore, we only bold the minimal runtime results.

\subsection{Performance on backsubstitution-based verifier}

CNN-Cert and DeepPoly are two state-of-the-art and popular backsubstitution-based verifiers. 
3DCertify, built atop DeepPoly, is a tight verifier for point cloud models. We conduct addtional experiments to compare Ti-Lin to CNN-Cert in Appendix due to the page limit.

In this experiment, we integrate Ti-Lin into DeepPoly and compare Ti-Lin to both DeepPoly and 3DCertify across various networks, including CNNs trained on MNIST and CIFAR-10 and PointNets trained on ModelNet40. The PointNets we use are from 3DCertify. $16$p, Natural, FGSM, and IBP represent the number point of inputs, naturally trained model, adversarial trained model by FGSM~\cite{DBLP:journals/corr/GoodfellowSS14,DBLP:conf/iclr/WongRK20}, and verified trained model by IBP~\cite{DBLP:conf/iccv/GowalDSBQUAMK19}, respectively.
The results are shown in Table~\ref{table-3dcertify}. The results demonstrate that Ti-Lin provides higher certified accuracy in all settings, improving by up to 39.4\% and 7.1\% over DeepPoly and 3DCertify on MNIST\_CNN\_8layers, respectively. As for time efficiency, 3DCertify computes the upper bound of every possible upper linear bound for the MaxPool function to choose the best upper linear bound, while Ti-Lin has provably neuron-wise tightest linear bounds and offers the best upper linear bound directly. Therefore, Ti-Lin is more efficient than 3DCertify with up to 5.3$\times$ speedup. Because both Ti-Lin and DeepPoly give the linear bounds directly, they share almost the same time efficiency.

\subsection{Performance on MN-based verifier}

MN-BaB is a state-of-the-art complete neural network verifier that builds on the tight multi-neuron constraints in PRIMA~\cite{muller2022prima}, and ERAN is the cutting-edge MN-based verifier. To evaluate the performance of Ti-Lin on the MN-based verifier, we compare Ti-Lin to both MN-BAB and ERAN in this experiment. We integrate Ti-Lin into ERAN to compare the performance of Ti-Lin to MN-BaB and ERAN. Concretely, we use the group-wise joint neuron abstractions~\cite{muller2022prima} for ReLU and MILP, and LP~\cite{singh2019boosting} solvers to analyze.

We compute the results of Ti-Lin built atop ERAN, which are shown in Table~\ref{table-eran}. The results show that  Ti-Lin verifies higher certified accuracy across all networks on different datasets with up to 39.4\% and 7.1\% improvement on MNIST\_CNN\_8layers, respectively.
In terms of efficiency, as ERAN uses MaxPool linear bounds of 3DCertify, which is time-consuming, Ti-Lin has less time cost in all cases than ERAN with up to 1.7$\times$ speedup on MNIST\_Conv\_MaxPool.
However, although Ti-Lin computes faster than MN-BaB on most networks trained on the MNIST dataset, both Ti-Lin and ERAN  are comparatively slower than MN-BaB on networks trained on the CIFAR-10 dataset, with a maximum slowdown of 2.6 $\times$ compared to MN-BaB on CIFAR\_Conv\_MaxPool. The reason is that although the timeout thresholds of these methods are all 180 seconds, the average runtime can vary across different frameworks. For example, ERAN may try its DeepPoly verifier first and then try using multi-neuron constraint to further tighten the output bounds, which may cause additional time consumption. For a fair comparison, we only compare the time efficiency of Ti-Lin to ERAN.

\begin{table*}[htbp]
\centering
\caption{The performance of Ti-Lin on $\alpha,\beta$-CROWN verifier (BaB-based). }
\huge
  \resizebox{0.9\textwidth}{!}{
      \begin{threeparttable}
\begin{tabular}{clrrrrrrrrrr}
\toprule
    & &   \multicolumn{3}{c}{{Certified Accuracy (\%)}}&\multicolumn{1}{c}{UB (\%)}&\multicolumn{3}{c}{{Timeout Rate(\%)} }&\multicolumn{3}{c}{{Average Runtime(min)} }
    \\\cmidrule(lr){10-12}\cmidrule(lr){7-9}\cmidrule(lr){3-5}\cmidrule(lr){6-6}
        \multicolumn{1}{c}{{Dataset }}&  \multicolumn{1}{c}{{Network }}& \multirow{1}{*}{{ $\alpha,\beta$-CROWN }}& \multicolumn{1}{c}{{MaxLin$^*$}} & \multicolumn{1}{c}{{Ti-Lin$^{**}$}} && \multirow{1}{*}{{$\alpha,\beta$-CROWN}}& \multicolumn{1}{c}{{MaxLin}}& \multicolumn{1}{c}{{Ti-Lin}}   & \multirow{1}{*}{{$\alpha,\beta$-CROWN}}& \multicolumn{1}{c}{{MaxLin}}& \multicolumn{1}{c}{{Ti-Lin}}  \\   
  \midrule
  \multirow{7}{*}{\shortstack{MNIST}}
  &Conv\_MaxPool&45.7 &60.5   & \textbf{66.7} &93.8    & 48.1     &33.3     & \textbf{27.2} & 1.7  &1.8 & \textbf{1.4} \\
  &Convnet\_MaxPool&4.2 &10.4    & \textbf{19.8} &71.9   & 21.9 &15.6& \textbf{6.3}  & 0.7  &0.5 & \textbf{0.3}\\
  &CNN, 4 layers$^{\dagger}$&79.8  &\textbf{92.6}  & \textbf{92.6}  &98.9  & 18.1  &\textbf{5.3} & \textbf{5.3}  & 0.8  &0.7 & \textbf{0.6}\\
  &CNN, 5 layers&9.1  &68.7   & \textbf{85.9}  &100.0  &90.9 &28.3 & \textbf{26.3} & 6.2  &3.2 & \textbf{2.1}\\
  &CNN, 6 layers&4.04  &71.7  & \textbf{73.7}  &99.0  & 94.9 &30.3 & \textbf{13.1} & 9.5  &3.8 & \textbf{3.3}\\
    &CNN, 7 layers&\textbf{0.0}   &\textbf{0.0}  & \textbf{0.0}   &100.0    & \textbf{98.9}  &\textbf{98.9} & \textbf{98.9} &11.8   &10.5 &\textbf{8.1} \\
  &CNN, 8 layers&0.0    &0.0    & \textbf{3.0}  &99.0    &98.0 &98.0& \textbf{94.9} & 14.2 &13.2 & \textbf{12.5} \\
\midrule
  \multirow{6}{*}{\shortstack{CIFAR-10}}
  &Conv\_MaxPool&2.1  &2.1  & \textbf{6.3}   &93.9  &58.3 &58.3 & \textbf{54.2} & 55.1 &50.2 & \textbf{11.9}\\
  &CNN, 4 layers&63.6  &74.1  & \textbf{90.9}  &100.0  & 36.4 &11.1 & \textbf{9.1}  & 2.3  &3.7 & \textbf{2.1}\\
  &CNN, 5 layers&0.0    &0.0   & \textbf{6.3}& 87.5& 81.3 &81.3 & \textbf{75.0}& 3.0 &4.9 &\textbf{2.5}\\
    &CNN, 6 layers&7.1  &57.1   & \textbf{85.7}  &100.0  & 92.9 &42.9 & \textbf{15.3} & 8.3  &6.6 & \textbf{5.9}\\
      &CNN, 7 layers&0.0   &\textbf{3.4}    & \textbf{3.4}  &72.7   & 86.2 &\textbf{82.8} & \textbf{82.8} & 8.3 &14.2  & \textbf{6.7}\\
  &CNN, 8 layers&\textbf{0.0}   &\textbf{0.0}& \textbf{0.0}   &91.7    & \textbf{83.3} &\textbf{83.3}& \textbf{83.3} & 12.5&11.6 & \textbf{11.2}\\
\bottomrule\\
\end{tabular}
    \begin{tablenotes}
        \item $^{\dagger}$ CNNs with 4-8 layers on MNIST and CIFAR-10, are initially saved in the ".h5" format and are converted into ".onnx" to be tested in this table. The transformation preserve the  same predict accuracy  but its robustness is affected~\cite{DBLP:conf/kbse/GuoCXMHLLZL19}. Thus, only the certified accuracy in the same table, rather than in different result tables, can effectively illustrate the performance of verifiers.
        \item $^*$ MaxLin, as proposed in its original paper, is designed to be integrated into various verification frameworks. In this work, we follow the approach in MaxLin to integrate it into $\alpha,\beta$-CROWN and evaluate its performance in the BaB-bound setting for ReLU.
        \item $^{**}$ In this table, Ti-Lin is also integrated into $\alpha,\beta$-CROWN for comparison.
    \end{tablenotes}
    \end{threeparttable}
    }
\label{table-alpha}
\end{table*}

\subsection{Performance on BaB-based verifier}
$\alpha,\beta$-CROWN is an efficient linear bound propagation framework and uses BaB for ReLU to tighten robustness results. 
To illustrate the superiority of Ti-Lin, which provides neuron-wise tightest linear bounds, over MaxLin, which offers block-wise tightest linear bounds, we integrate both Ti-Lin and MaxLin into the $\alpha,\beta$-CROWN framework. Concretely, in addition to the default settings of the $\alpha,\beta$-CROWN framework, we allocate more computational time for the BaB procedure in the ReLU layers and set the timeout for BaB to 3600 seconds, ensuring the verifier can explore the feasible region for ReLU more thoroughly during the verification process.

The results, shown in Table~\ref{table-alpha}, demonstrate that Ti-Lin can improve the certified accuracy by up to 78.6\% compared to $\alpha,\beta$-CROWN, and by 28.6\% compared to MaxLin implemented in $\alpha,\beta$-CROWN on the CIFAR\_CNN\_6layer model.
This reveals that verifiers with the neuron-wise linear bounds for MaxPool can greatly reduce the overestimation and compute higher verified accuracy. 
As for efficiency, our method can accelerate the computation process in all cases in Table~\ref{table-alpha}. The speedup over $\alpha,\beta$-CROWN is due to its adaptive linear bounds for MaxPool, which are gradually optimized during the propagation process, resulting in more time consumption. While the speedup over MaxLin is attributed to Ti-Lin's higher precision when using BaB for ReLU, allowing Ti-Lin to handle fewer time-out samples, which leads to faster computation.

\subsection{Additional experiments}
We conduct additional experiments to further demonstrate the superiority 
 of Ti-Lin.
The detailed experiment settings are in the Appendix, and we perform the following experiments:
(I) We conduct extensive experiments by comparing Ti-Lin to CNN-Cert, a backsubstitution-based verifier, by integrating our neuron-wise tightest linear approximation into CNN-Cert.
(II) We compare Ti-Lin to OSIP to illustrate the superiority of tight linear approximation for MaxPool over the MaxPool2ReLU transformation.
(III) We analyze the global lower bounds of the last layer computed by Ti-Lin and other methods, including DeepPoly, 3DCertify,  optimized linear bounds, used in $\alpha,\beta$-CROWN, and MaxLin to further illustrate the advantages of the neuron-wise tightness over other linear bounds. 
(IV) To illustrate the difference between MaxLin and Ti-Lin, we introduce \textcolor{red}{Hybrid-Lin} in the Appendix, Hybrid-Lin applies MaxLin when the ReLU upper linear bound is $u(x)=\frac{u}{u-l}(x-l)$ and Ti-Lin when the ReLU upper linear bound is $u(x)=0$ or $u(x)=x$. We compare the CROWN and $\alpha$-CROWN results of Ti-Lin, MaxLin, and Hybrid-Lin, demonstrating that \textbf{\textit{rather than competing, MaxLin and Ti-Lin complement each other in achieving optimal bounding precision.}}

\section{Conclusion}
\label{section7}
In this paper, we propose Ti-Lin, a robustness verifier for MaxPool-based CNNs.
 We are the first to propose the neuron-wise tightest linear bounds for the MaxPool function. 
 We implemented Ti-Lin in various networks on the MNIST, CIFAR-10, Tiny ImageNet, and ModelNet40 datasets. Experimental results show the effectiveness of the neuron-wise tightest techniques with up to 78.6\% improvement to the certified accuracy over the state-of-the-art linear approximation methods with the same time consumption as the fastest tools. The results also show that transforming MaxPool to ReLU can lead to more time consumption and more coarse verification results. Moreover, as Ti-Lin is simple to be integrated into other verifiers, the results after integration strongly show that Ti-Lin  reduce the overestimation and compute tighter certified robustness results.

 \vspace{-10pt}
\paragraph{Acknowledgements.} The authors would like to thank the anonymous reviewers for their insightful comments. This work is partially supported by the National Natural Science Foundation of China (U24A20337, 62372228), the Shenzhen-Hong Kong-Macau Technology Research Programme (Type C) (Grant No. SGDX20230821091559018), and the Fundamental Research Funds for the Central Universities (14380029).

{
    \small
    \bibliographystyle{ieeenat_fullname}
    \bibliography{reference}
}

\clearpage
\newpage
 \onecolumn
\section{Appendix}
\label{section8}

\subsection{Proof of Theorem~\ref{theoremMaxPool}}
\label{proofmaxpool}

In this subsection, we prove the soundness of our linear bounds in Theorem~\ref{theoremMaxPool}.

\begin{proof}
Define $\boldsymbol{m}=\frac{\boldsymbol{u}+\boldsymbol{l}}{2}=(m_1,\cdots,m_n)\in\mathbb{R}^n$.

\textbf{Upper linear bound:}

Case 1: When  $l_i=l_{max}\wedge l_{max}\geq u_j$, we have$f(x_1,\cdots,x_n)=x_i$ and  $u(x_1,\cdots,x_n)=x_i-l_i+l_i=x_i$.
Then, we have$u(\boldsymbol{x})\geq f(\boldsymbol{x})$, that is, the upper linear bound of case 1 is sound.

Case 2: When  $l_i=l_{max}, u_j>l_i\geq u_k$, we have $f(x_1,\cdots,x_n)=max(x_i,x_j)$ and $u(x_1,\cdots,x_n)=x_i-l_i+\frac{u_j-l_i}{u_j-l_j}(x_j-l_j)+l_i$. 

If $ f(x_1,\cdots,x_n)=x_i$, $$\begin{aligned}u(x_1,\cdots,x_n)-x_i=&x_i-l_i+\frac{u_j-l_i}{u_j-l_j}(x_j-l_j)+l_i-x_i\\
=&\frac{u_j-l_i}{u_j-l_j}(x_j-l_i)\\&\geq 0
\end{aligned}$$

If $ f(x_1,\cdots,x_n)=x_j$, $$\begin{aligned}u(x_1,\cdots,x_n)-x_j=&x_i-l_i+\frac{u_j-l_i}{u_j-l_j}(x_j-l_j)+l_i-x_j\\
=&(x_i-l_i)+\frac{u_j-l_i}{u_j-l_j}(x_j-l_j)+l_j-x_j-l_j+l_i\\
=&(x_i-l_i)+\frac{l_j-l_i}{u_j-l_j}(x_j-l_j)-l_j+l_i\\
=&(x_i-l_i)+\frac{u_j-x_j}{u_j-l_j}(l_i-l_j)\\\geq& 0
\end{aligned}$$

Then, we have $u(\boldsymbol{x})\geq f(\boldsymbol{x})$, that is, the upper linear bound of case 2 is sound.

Case 3: When $l_i=l_{max}\wedge l_i\neq l_{max}\wedge u_j\geq l_j\geq u_k$, we have $f(x_1,\cdots,x_n)=max(x_i,x_j)$ and $u(x_1,\cdots,x_n)=\frac{u_i-l_j}{u_i-l_i}x_i+x_j+l_j$. 

If $ f(x_1,\cdots,x_n)=x_i$, $$\begin{aligned}u(x_1,\cdots,x_n)-x_i=&\frac{u_i-l_j}{u_i-l_i}(x_i-l_i)+(x_j-l_j)+l_j-l_i+l_i-x_i\\
=&(x_i-l_i)(\frac{u_i-l_j}{u_i-l_i}-1)+(x_j-l_j)+l_j-l_i\\
=&(x_i-l_i)\frac{l_i-l_j}{u_i-l_i}+(x_j-l_j)+l_j-l_i\\
=&(l_j-l_i)(1-\frac{x_i-l_i}{u_i-l_i})+(x_j-l_j)\\
=&(l_j-l_i)\frac{u_i-x_i}{u_i-l_i}+(x_j-l_j)\\\geq&0
\end{aligned}$$

 If $ f(x_1,\cdots,x_n)=x_j$, $$\begin{aligned}u(x_1,\cdots,x_n)-x_j=&\frac{u_i-l_j}{u_i-l_i}(x_i-l_i)+(x_j-l_j)+l_j-x_j\\
=&\frac{u_i-l_j}{u_i-l_i}(x_i-l_i)\\\geq&0
\end{aligned}$$

Then, we have $u(\boldsymbol{x})\geq f(\boldsymbol{x})$, that is, the upper linear bound of case 3 is sound.

Case 4: First, we prove that if $\boldsymbol{u}$ and $\boldsymbol{l}$ do not satisfy case 1,2, and 3, then $u_k>max\{l_i,l_j\}$.

We prove this by contradiction.

We assume that $u_k\leq max\{l_i,l_j\}$. Then, as $u_k\geq l_k$ and $u_k\geq max_{s\neq i,j,k}\{u_p\}$, we have $u_k\geq max_{p\neq i,j}\{l_p\}$.

And we have $l_{max}=max\{l_1,\cdots,l_n\}=max\{l_i,l_j,u_k\}=max\{l_i,l_j\}$.

If $l_{max}=l_i$, then we have $l_i<u_j \wedge (l_i< u_k \vee l_i\geq u_j)$, that is, $l_i<u_j \wedge l_i< u_k$. It contradicts $u_k\leq max\{l_i,l_j\}$.

if $l_{max}=l_j \wedge l_{max}\neq l_i$, then we have $l_j>u_j\vee l_j<u_k$, that is $l_j<u_k$.  It contradicts $u_k\leq max\{l_i,l_j\}$. 

Therefore, in case 4,  $u_k>max\{l_i,l_j\}$.

Here, we prove the upper bound in case 4 is sound.

$f(x_1,\cdots,x_n)=max(x_1,\cdots,x_n),\forall (x_1,\cdots,x_n)$. If $ f(x_1,\cdots,x_n)=x_i$, $$\begin{aligned}u(x_1,\cdots,x_n)-x_i=&\frac{u_i-u_k}{u_i-l_i}(x_i-l_i)+\frac{u_j-u_k}{u_j-l_j}(x_j-l_j)+u_k-x_i\\
=&\frac{u_i-u_k}{u_i-u_k}(x_i-l_i)+\frac{u_j-u_k}{u_j-l_j}(x_j-l_j)+u_k-l_i+l_i-x_i\\
=&(x_i-l_i)(\frac{u_i-u_k}{u_i-l_i}-1)+\frac{u_j-u_k}{u_j-l_j}(x_j-l_j)+u_k-l_i\\
=&(u_k-l_i)\frac{u_i-x_i}{u_i-l_i}+\frac{u_j-u_k}{u_j-l_j}(x_j-l_j)\\\geq& 0
\end{aligned}$$

 If $ f(x_1,\cdots,x_n)=x_j$, the proof is the same as above.
 $$\begin{aligned}u(x_1,\cdots,x_n)-x_j=&\frac{u_i-u_k}{u_i-l_i}(x_i-l_i)+\frac{u_j-u_k}{u_j-l_j}(x_j-l_j)+u_k-x_j\\&
=\frac{u_i-u_k}{u_i-u_k}(x_i-l_i)+\frac{u_j-u_k}{u_j-l_j}(x_j-l_j)+u_k-l_j+l_j-x_j\\&
=\frac{u_i-u_k}{u_i-l_i}(x_i-l_i)+(\frac{u_j-u_k}{u_j-l_j}-1)(x_j-l_j)+u_k-l_j\\&
=\frac{u_i-u_k}{u_i-l_i}(x_i-l_i)+(1-\frac{x_j-l_j}{u_j-l_j})(u_k-l_j)\\&
=\frac{u_i-u_k}{u_i-l_i}(x_i-l_i)+\frac{u_j-x_j}{u_j-l_j}(u_k-l_j)\\&\geq 0
\end{aligned}$$
 
 If $ f(x_1,\cdots,x_n)=x_k$, 
 $$\begin{aligned}u(x_1,\cdots,x_n)-x_j&=\frac{u_i-u_k}{u_i-l_i}(x_i-l_i)+\frac{u_j-u_k}{u_j-l_j}(x_j-l_j)+(u_k-x_k)\\&\geq 0
\end{aligned}$$ 
 
  If $ f(x_1,\cdots,x_n)=x_l,l\neq i,j,k$,
 $$\begin{aligned}&u(x_1,\cdots,x_n)-x_l\\=&\frac{u_i-u_k}{u_i-l_i}(x_i-l_i)+\frac{u_j-u_k}{u_j-l_j}(x_j-l_j)+u_k-x_l\\
 =&\frac{u_i-u_k}{u_i-l_i}(x_i-l_i)+\frac{u_j-u_k}{u_j-l_j}(x_j-l_j)+(u_k-u_l)+(u_l-x_l)\\\geq& 0
\end{aligned}$$ 

Then, we have $u(\boldsymbol{x})\geq f(\boldsymbol{x})$, that is, the upper linear bound of case 4 is sound.

\textbf{Lower linear bound:}

$l(x_1,\cdots,x_n)=x_j=argmax_i m_i$, and  $\forall (x_1,\cdots,x_n)\in \times_{i=1}^n[l_i,u_i]$, 
$$\begin{aligned}
    f(x_1,\cdots,x_n)&=max(x_1,\cdots,x_n)\\&\geq x_j\\&=l(x_1,\cdots,x_n)
\end{aligned}$$
Then, we have $l(\boldsymbol{x})\leq f(\boldsymbol{x})$, that is the lower linear bound is sound.

This completes the proof.
\end{proof}

\subsection{Proof of Theorem~\ref{theorem-generalbound}}
\label{section-general}

In this subsection, we prove Theorem~\ref{theorem-generalbound}.

\begin{proof}
First, $u(\boldsymbol{x})$ and $l(\boldsymbol{x})$ are upper bound and lower bound for $f(\boldsymbol{x})$. Therefore, $u(\boldsymbol{x})\geq l(\boldsymbol{x}), \forall \boldsymbol{x}\in[\boldsymbol{l},\boldsymbol{u}]$. Then, we have$$u(\boldsymbol{x})-l(\boldsymbol{x})=(u(\boldsymbol{x})-f(\boldsymbol{x}))+(f(\boldsymbol{x})-l(\boldsymbol{x}))$$

$(u(\boldsymbol{x})-f(\boldsymbol{x}))$ and $(f(\boldsymbol{x})-l(\boldsymbol{x}))$ are not smaller than 0, when $ \forall \boldsymbol{x}\in[\boldsymbol{l},\boldsymbol{u}]$.

Therefore, minimizing $\iint_{[\boldsymbol{l},\boldsymbol{u}]}(u(\boldsymbol{x})-l(\boldsymbol{x}))d\boldsymbol{x}$ is equivalent to minimizing both $\iint_{[\boldsymbol{l},\boldsymbol{u}]}(u(\boldsymbol{x})-f(\boldsymbol{x}))d\boldsymbol{x}$ and $\iint_{[\boldsymbol{l},\boldsymbol{u}]}(f(\boldsymbol{x})-l(\boldsymbol{x}))d\boldsymbol{x}$. The value of $\iint_{[\boldsymbol{l},\boldsymbol{u}]}f(\boldsymbol{x})d\boldsymbol{x}$ is constant. Thus, it is also  equivalent to minimizing $\iint_{[\boldsymbol{l},\boldsymbol{u}]}u(\boldsymbol{x})d\boldsymbol{x}$ and $\iint_{[\boldsymbol{l},\boldsymbol{u}]}(-l(\boldsymbol{x}))d\boldsymbol{x}$.
Further, we define that lower linear bound $l(\cdot)$ is the neuron-wise tightest when $-l(\boldsymbol{m})$ reaches the minimum, and upper linear bound $u(\cdot)$ is the neuron-wise tightest when $u(\boldsymbol{m})$ reaches the minimum.

Define $\boldsymbol{x}=(x_1,\cdots,x_{n})$ and $[n]=\{1,2,\cdots,n\}$.
Because $u(\boldsymbol{x})$ is a linear combination of $x_i,i\in[n]$. Without loss of generality, we assume $u(\boldsymbol{x})=\sum_{i\in[n]}a_{u,i} x_i+b_{u}$. Then,
\begin{equation}
\begin{aligned}
\iint_{(x_1,\cdots,x_{n})\in[\boldsymbol{l},\boldsymbol{u}]}u(x_1,\cdots,x_{n})d\boldsymbol{x}
=&\iint_{(x_1,\cdots,x_{n})\in[\boldsymbol{l},\boldsymbol{u}]}(\sum_{i\in[n]}a_{u,i} x_i+b_{u})d\boldsymbol{x}\\
=&\sum_{i=1}^n\frac{a_{u,i}}{2}((u_i)^2-(l_i)^2)+b_{u}(u_i-l_i)\\
=&\prod_{i=1}^n(u_i-l_i)(\sum_{i\in[n]}a_{u,i}\frac{u_i+l_i}{2}+b_{u})\\
=&\prod_{i=1}^n(u_i-l_i)u(\boldsymbol{m})\\
\end{aligned}\nonumber
\end{equation}
    
where $\boldsymbol{m}=(\frac{u_1+l_1}{2},\cdots,\frac{u_{n}+l_{n}}{2})$. As $u_i,l_i,i\in[n]$ are constant,  the minimize target has been transformed into minimizing $u(\boldsymbol{m})$.

Symmetric to the above proof,  minimizing $\iint_{(x_1,\cdots,x_{n})\in[\boldsymbol{l},\boldsymbol{u}]}u(x_1,\cdots,x_{n})d\boldsymbol{x}$ is equivalent to minimize $-l(\boldsymbol{m})$.

This completes the proof.
\end{proof}

\subsection{Proof of Theorem~\ref{tightest}}
\label{section-tightest}
In this subsection, we prove Theorem~\ref{tightest}, that is, our linear bounds in Theorem~\ref{theoremMaxPool} are the neuron-wise tightest.

\begin{proof}
Define $\boldsymbol{m}=\frac{\boldsymbol{u}+\boldsymbol{l}}{2}=(m_1,\cdots,m_n)\in\mathbb{R}^n$.

\textbf{Upper linear bound:}

Case 1: When  $l_i=l_{max}\wedge l_{max}\geq u_j$, we have$f(x_1,\cdots,x_n)=x_i$ and  $u(x_1,\cdots,x_n)=x_i-l_i+l_i=x_i$.
As $u(m)=f(m)\leq u'(\boldsymbol{m}),\forall u'\in\mathcal{U}$, the upper bound is the neuron-wise tightest.

Case 2: When  $l_i=l_{max}, u_j>l_i\geq u_k$, we have $f(x_1,\cdots,x_n)=max(x_i,x_j)$ and $u(x_1,\cdots,x_n)=x_i-l_i+\frac{u_j-l_i}{u_j-l_j}(x_j-l_j)+l_i$. 

Because
$$
\begin{aligned}&u(u_1,\cdots,u_{i-1},l_i,u_{i+1},\cdots,u_{j-1},u_j,u_{j+1},\cdots,u_n)\\=&l_i-l_i+\frac{u_j-l_i}{u_j-l_j}(u_j-l_j)+l_i\\
=&u_j\\
=&f(u_1,\cdots,u_{i-1},l_i,u_{i+1},\cdots,u_{j-1},u_j,u_{j+1},\cdots,u_n)
\end{aligned}$$

and 

$$
\begin{aligned}&u(l_1,\cdots,l_{i-1},u_i,l_{i+1},\cdots,l_{j-1},l_j,l_{j+1},\cdots,l_n)\\=&u_i-l_i+\frac{u_j-l_i}{u_j-l_j}(l_j-l_j)+l_i\\
=&u_i\\
=&f(l_1,\cdots,l_{i-1},u_i,l_{i+1},\cdots,l_{j-1},l_j,l_{j+1},\cdots,l_n)\\
\end{aligned}$$ 

We notice that 

$\boldsymbol{a}:=(u_1,\cdots,u_{i-1},l_i,u_{i+1},\cdots,u_n)$ and $\boldsymbol{b}:=(l_1,\cdots,l_{i-1},u_i,l_{i+1},\cdots,l_n)$ are the space diagonal of $\times_{i=1}^n[l_i,u_i]$, and 

$\boldsymbol{m}=\frac{1}{2}(\boldsymbol{a} +\boldsymbol{b})$.  Then, as $u(\boldsymbol{x})$ is linear, we have $f(\boldsymbol{a})+f(\boldsymbol{b})=f(\boldsymbol{a}+\boldsymbol{b})$.

Then, we have
\begin{equation}
\begin{aligned}
      \forall u'\in\mathcal{U}, u(\boldsymbol{m})&=u(\frac{1}{2}(\boldsymbol{a}+\boldsymbol{b}))\\
      &=\frac{1}{2}(u(\boldsymbol{a})+u(\boldsymbol{b}))\\
     &=\frac{1}{2}(f(\boldsymbol{a})+f(\boldsymbol{b}))\\
     &\leq\frac{1}{2}(u'(\boldsymbol{a})+u'(\boldsymbol{b}))\\
     &=(u'(\frac{1}{2}(\boldsymbol{a}+\boldsymbol{b})))\\
     &=u'(\boldsymbol{m})
\end{aligned}
\nonumber
\end{equation}

Therefore, the plane is the neuron-wise tightest  upper linear bounding plane.

Case 3: When $l_i=l_{max}\wedge l_i\neq l_{max}\wedge u_j\geq l_j\geq u_k$, we have $f(x_1,\cdots,x_n)=max(x_i,x_j)$ and $u(x_1,\cdots,x_n)=\frac{u_i-l_j}{u_i-l_i}x_i+x_j+l_j$.

Because
$$
\begin{aligned}&u(u_1,\cdots,u_{i-1},u_i,u_{i+1},\cdots,u_{j-1},l_j,u_{j+1},\cdots,u_n)\\=&\frac{u_i-l_j}{u_i-l_i}(u_i-l_i)+(l_j-l_j)+l_j\\
=&f(u_1,\cdots,u_{i-1},u_i,u_{i+1},\cdots,u_{j-1},l_j,u_{j+1},\cdots,u_n)
\end{aligned}$$

and 

$$
\begin{aligned}&u(l_1,\cdots,l_{i-1},l_i,l_{i+1},\cdots,l_{j-1},u_j,l_{j+1},\cdots,l_n)\\=&\frac{u_i-l_j}{u_i-l_i}(l_i-l_i)+(u_j-l_j)+l_j\\
=&f(l_1,\cdots,l_{i-1},l_i,l_{i+1},\cdots,l_{j-1},u_j,l_{j+1},\cdots,l_n)\\
\end{aligned}$$ 

We notice that 

$\boldsymbol{a}:=(u_1,\cdots,u_{j-1},l_j,u_{j+1},\cdots,u_n)$ and $\boldsymbol{b}:=(l_1,\cdots,l_{j-1},u_j,l_{j+1},\cdots,l_n)$ are the space diagonal of $\times_{i=1}^n[l_i,u_i]$, and 

$\boldsymbol{m}=\frac{1}{2}(\boldsymbol{a} +\boldsymbol{b})$. Similar to the proof in case 2, we can prove that the plane is the neuron-wise tightest linear upper bounding  plane.

Case 4: First, as proved in Section~\ref{proofmaxpool},  $u_k>max\{l_i,l_j\}$ in case 4.

Because 
$$\begin{aligned}
&u(u_1,\cdots,u_{i-1},u_i,u_{i+1},\cdots,u_{j-1},l_j,u_{j+1},\cdots,u_n)\\=&\frac{u_i-u_k}{u_i-l_i}(u_i-l_i)+\frac{u_j-u_k}{u_j-l_j}(l_j-l_j)+u_k\\
=&u_i-u_k+u_k\\
=&f(u_1,\cdots,u_{i-1},u_i,u_{i+1},\cdots,u_{j-1},l_j,u_{j+1},\cdots,u_n)\\
\end{aligned}$$

and 

$$\begin{aligned}
&u(l_1,\cdots,l_{i-1},l_i,l_{i+1},\cdots,l_{j-1},u_j,l_{j+1},\cdots,l_n)\\=&\frac{u_i-u_k}{u_i-l_i}(l_i-l_i)+\frac{u_j-u_k}{u_j-l_j}(u_j-l_j)+u_k\\
=&u_j-u_k+u_k\\
=&f(l_1,\cdots,l_{i-1},l_i,l_{i+1},\cdots,l_{j-1},u_j,l_{j+1},\cdots,l_n)\\
\end{aligned}$$ 
As
$$\begin{aligned}\boldsymbol{m}=&\frac{1}{2}(u_1,\cdots,u_{i-1},u_i,u_{i+1},\cdots,u_{j-1},l_j,u_{j+1},\cdots,u_n)\\+&\frac{1}{2}(l_1,\cdots,l_{i-1},l_i,l_{i+1},\cdots,l_{j-1},u_j,l_{j+1},\cdots,l_n)\end{aligned}$$
We notice that 

$\boldsymbol{a}:=(l_1,\cdots,l_{j-1},u_j,l_{j+1},\cdots,l_n)$ 
and 
$\boldsymbol{b}:=(u_1,\cdots,u_{j-1},l_j,u_{j+1},\cdots,u_n))$ are the space diagonal of $\times_{i=1}^n[l_i,u_i]$, and 
$\boldsymbol{m}=\frac{1}{2}(\boldsymbol{a} +\boldsymbol{b})$. 
Similar to the proof in case 2, we can prove that the plane is the neuron-wise tightest linear upper bounding plane.

\textbf{Lower linear bound:}

$l(x_1,\cdots,x_n)=x_j=argmax_i m_i$, and  $\forall (x_1,\cdots,x_n)\in \times_{i=1}^n[l_i,u_i]$, 

 $l(\boldsymbol{m})=f(m_1,\cdots,m_n)\geq l'(\boldsymbol{m}),\forall l\in\mathcal{L}$, hence, $l(x_1,\cdots,x_n)$ is the neuron-wise tightest lower bounding plane.

This completes the proof.
\end{proof}

 \vspace{-0.6cm}
 \renewcommand\arraystretch{0.8}
\begin{table}[htbp]
\centering
\tiny
\caption{The additional experimental setup and source  of neural networks used in experiments.}
\resizebox{0.9\columnwidth}{!}{
\begin{tabular}{clrrrrr}
\toprule {Dataset  }  &  {Network} & \multicolumn{1}{r}{ {\#Nodes}}& \multicolumn{1}{r}{ {Accuracy}} & \multicolumn{1}{r}{ {\#Properties}} &\multicolumn{1}{r}{{${\epsilon}$}}&\multicolumn{1}{r}{{Source}} \\\midrule
\multirow{7}{*}{MNIST}&Conv\_MaxPool&14592& 81.9&81&2/255&ERAN \\\cmidrule(rl){2-7}
&Convnet\_MaxPool &25274&98.8&96&10/255&\multirow{1}{*}{Verivital}\\\cmidrule(rl){2-7}
&CNN, 4 layers &36584&99.0&94&3/255& \multirow{5}{*}{CNN-Cert}\\
&CNN, 5 layers &52872&99.1&99&2/255& \\
&CNN, 6 layers &56392&99.1&99&2/255& \\
&CNN, 7 layers &56592&99.1&91&2/255& \\
&CNN, 8 layers &56912&99.3&99&2/255& \\
\midrule
\multirow{6}{*}{CIFAR-10}&Conv\_MaxPool&57020&44.6&48& 0.001&ERAN\\\cmidrule(rl){2-7}
&CNN, 4 layers &49320&71.3&27&0.001&\multirow{5}{*}{CNN-Cert}\\
&CNN, 5 layers &71880&71.1&16&0.001&\\
&CNN, 6 layers &77576&73.8&14&0.001&\\
&CNN, 7 layers &77776&75.6&29&0.001&\\
&CNN, 8 layers & 78416&68.1&12&0.001&\\\midrule
\multirow{5}{*}{ModelNet40}
&16p\_Natural &26200&76.8&59&0.005&\multirow{5}{*}{3DCertify}\\
&32p\_Natural &49800&83.2&62&0.005&\\
&64p\_Natural &97000&85.7&64&0.005&\\
&64p\_FGSM &97000&86.0&66&0.01&\\
&64p\_IBP &97000&78.1&60&0.01&\\
\bottomrule
\end{tabular}}
\label{table-network}
\end{table}

\renewcommand\arraystretch{1.0}
\begin{table*}[t]
    \centering
    \tiny
    \caption{The performance of Ti-Lin on CNN-Cert (backsubstitution-based).}
        \huge
    \resizebox{0.9\textwidth}{!}{
    \begin{tabular}{cccrrrrr}
    \toprule
       & &&\multicolumn{2}{c}{{Certified Bounds($\boldsymbol{10^{-5}}$)}}  &\multicolumn{1}{c}{{Bound Impr.(\%)}} &\multicolumn{2}{c}{{Average Runtime(min)} }\\\cmidrule(lr){4-5}\cmidrule(lr){6-6}\cmidrule(lr){7-8}
      {Dataset }  &  {Network  } & $l_p$   &\multirow{1}{*}{{CNN-Cert}}& \multicolumn{1}{c}{{Ti-Lin, on CNN-Cert}}&\multirow{1}{*}{{vs. CNN-Cert }}& \multirow{1}{*}{{CNN-Cert }  }&\multicolumn{1}{c}{   Ti-Lin, on CNN-Cert    }     \\
      \midrule
    \multirow{27}{*}{MNIST }& CNN&$l_\infty$&1318&\textbf{1837}&39.4&1.8&1.7\\
      &4 layers&$l_2$&4427&\textbf{6478}&46.3&1.4&1.4\\
    &36584 nodes & $l_1$&8544&\textbf{12642}&48.0&1.4&1.4	\\\cmidrule(rl){2-8}
    &CNN & $l_\infty$&1288&\textbf{1817}&41.0&8.4&8.8\\
           & 5 layers     & $l_2$&5164&\textbf{7359}&42.5&11.9&9.2\\
    &  52872 nodes & $l_1$&10147&\textbf{14292}&40.9&10.8&9.5	 \\\cmidrule(rl){2-8}
      &CNN & $l_\infty$&1025&\textbf{1382}&34.8&20.5&20.9\\
         & 6 layers   & $l_2$&3954&\textbf{5409}&36.8&20.6&20.4\\
    &56392 nodes & $l_1$& 7708              &\textbf{10455} &35.6&  20.6&	20.0\\\cmidrule(rl){2-8}
     & CNN& $l_\infty$&  647                  &\textbf{930}    &43.7&     24.7	&24.6    \\
        &  7 layers     & $l_2$&2733                     & \textbf{4022} &47.2&  25.1&	23.8        \\
    &56592 nodes& $l_1$&   5443                     & \textbf{8002}            &47.0&  22.9	&22.9	\\\cmidrule(rl){2-8}
    &CNN & $l_\infty$& 847                     & \textbf{1221}   &44.2&         26.5&	26.7	\\
         & 8 layers     & $l_2$&  3751                 &\textbf{5320}       &41.8&    25.0&	24.9	  \\
     & 56912 nodes & $l_1$& 7515                     & \textbf{10655}        &41.8&      23.7	&24.2   \\\cmidrule(rl){2-8}
    &LeNet\_ReLU   & $l_\infty$&  1204&\textbf{1864}&54.8&0.2	&0.2 \\
       & 3 layers    & $l_2$& 6534&\textbf{10862}&66.2&0.2&	0.2\\
    &8080 nodes& $l_1$&  17937&\textbf{30305}&69.0&0.2	&0.2\\\cmidrule(rl){2-8}
     &LeNet\_Sigmoid   & $l_\infty$& 1684 &\textbf{2042} &21.3& 0.3&	0.3\\
         & 3 layers   & $l_2$&9926&\textbf{12369}&24.6&0.3&	0.3	\\
    &8080 nodes & $l_1$&26937 &\textbf{33384} &23.9&0.3	&0.3 \\\cmidrule(rl){2-8}
     & LeNet\_Tanh  &$l_\infty$&613 &\textbf{817}&33.3&0.3	&0.3\\
      &3 layers    & $l_2$&3462&\textbf{4916}&42.0& 0.3&	0.3	 \\
     &8080 nodes& $l_1$&9566&\textbf{13672} &42.9&0.3	&0.3	\\\cmidrule(rl){2-8}
     &  LeNet\_Atan   & $l_\infty$&617&\textbf{836} &35.5& 0.3&	0.3 \\
     & 3 layers  & $l_2$&3514&\textbf{5010} &42.6& 0.3&	0.3	  \\
    &8080 nodes & $l_1$&9330&\textbf{13345} &43.0& 0.3	&0.3\\\midrule 
    \multirow{15}{*}{CIFAR-10}&CNN & $l_\infty$&108                     &\textbf{129}      &19.4&3.1&	2.9	\\
           &4 layers    & $l_2$&   {751}&\textbf{1038}     &38.2& 2.5&	2.5\\
    &49320 nodes   & $l_1$&   2127                     & \textbf{3029}&42.4&2.5&	2.5	\\\cmidrule(rl){2-8}
     &CNN & $l_\infty$& 115                     & \textbf{146} &27.0&13.1	&13.0	\\
     &     5 layers    & $l_2$&   953                     &\textbf{1342}&40.8& 12.4&	12.7\\
    &71880 nodes     & $l_1$&    {2850}  & \textbf{4087}               &43.4&12.3	&12.6\\\cmidrule(rl){2-8}
    &CNN & $l_\infty$& 99                     & \textbf{120}     &21.2& 28.6&	28.6	\\
          &6 layers    & $l_2$&   830                      &\textbf{1078}            &29.9&27.6	&27.9\\
     & 77576 nodes  & $l_1$&  2387                     & \textbf{3174}         &33.0&27.7&	27.4 \\\cmidrule(rl){2-8}
     &CNN  & $l_\infty$& 66                     & \textbf{83}    &25.8& 33.4	&33.3 \\
          & 7 layers    & $l_2$& 573                  &\textbf{773}          &34.9&  32.5&	32.8	\\
     & 77776 nodes  & $l_1$& 1673              &\textbf{2303}             &37.7&  33.6	&32.6	 \\\cmidrule(rl){2-8}
     &CNN & $l_\infty$& 56                     &    \textbf{70}  &25.0&   36.9&	37.5	   \\
         & 8 layers    & $l_2$&   536                     & \textbf{705}   &31.5& 37.5	&36.6	\\
    &  78416 nodes & $l_1$& 1609                     & \textbf{2160}     &34.2&36.9	&37.0 \\\midrule  
  \multirow{3}{*}{Tiny ImageNet}  &CNN& $l_\infty$ &77 &\textbf{123}&59.7 &184.9&184.0\\
       &      7 layers  & $l_2$&580&\textbf{939}&61.9&184.4&183.3\\
 & 703512 nodes & $l_1$  &1747&\textbf{2875}&64.6  &193.6 &183.9
\\  \bottomrule
    \end{tabular}}
    \label{table-cnncert}
    \end{table*}
\begin{table*}[!t]

\caption{The performance of Ti-Lin and MaxPool2ReLU on $\alpha,\beta$-CROWN verifier.}
\centering
\tiny
    \resizebox{0.9\textwidth}{!}{
\begin{tabular}{lrrrrrr}
\toprule
&\multicolumn{1}{l}{}&\multicolumn{2}{c}{{Certified accuracy(\%) }}&\multicolumn{2}{c}{{Avg. time of safe instances(s) }}&\multicolumn{1}{c}{{Speedup}}\\\cmidrule(lr){5-6}\cmidrule(lr){3-4}\cmidrule(lr){7-7}
   \multicolumn{1}{l}{{Network}}&\multicolumn{1}{l}{{Radius}}&\multicolumn{1}{c}{{MaxPool2ReLU}}&\multicolumn{1}{c}{{Ti-Lin}}&\multicolumn{1}{c}{{MaxPool2ReLU}}&\multicolumn{1}{c}{{Ti-Lin}}&\multicolumn{1}{c}{{vs. MaxPool2ReLU}}\\\midrule
\multirow{3}{*}{\shortstack[l]{MNIST\_Conv\_MaxPool}}
&0.008&56.8&\textbf{65.4}&32.5&\textbf{17.3}&1.9\\
&0.009&49.4&\textbf{60.5}&46.3&\textbf{25.2}&1.8\\
&0.010&40.7&\textbf{54.3}&65.9&\textbf{32.5}&2.0\\\midrule
\multirow{3}{*}{\shortstack[l]{ConvNet\_MaxPool}}&0.020&60.0&\textbf{80.0}&191.0&\textbf{0.6}&318.3\\
&0.030&15.0&\textbf{50.0}&377.5&\textbf{37.5}&10.1\\
&0.040&0.0&\textbf{20.0}&-&\textbf{44.2}&-\\
\midrule

&\multicolumn{1}{l}{}&\multicolumn{2}{c}{{Timeout rate(\%)}}&\multicolumn{2}{c}{{Avg. time of  all instances(s) }}&\multicolumn{1}{c}{{Speedup}}\\\cmidrule(lr){5-6}\cmidrule(lr){3-4}\cmidrule(lr){7-7}
   \multicolumn{1}{l}{{Network}}&\multicolumn{1}{l}{{Radius}}&\multicolumn{1}{c}{{MaxPool2ReLU}}&\multicolumn{1}{c}{{Ti-Lin}}&\multicolumn{1}{c}{{MaxPool2ReLU}}&\multicolumn{1}{c}{{Ti-Lin}}&\multicolumn{1}{c}{{vs. MaxPool2ReLU}}\\\midrule
\multirow{3}{*}{\shortstack[l]{MNIST\_Conv\_MaxPool}}
&0.008&37.0&\textbf{28.4}&494.9&\textbf{65.5}&7.6\\
&0.009&43.2&\textbf{32.1}&667.8&\textbf{77.4}&8.6\\
&0.010&50.6&\textbf{37.0}&770.1&\textbf{88.7}&{8.7}\\\midrule

\multirow{3}{*}{\shortstack[l]{ConvNet\_MaxPool}}
&0.020&20.0&\textbf{0.0}&294.6&\textbf{0.7}&{420.9}\\
&0.030&60.0&\textbf{20.0}&549.5&\textbf{104.7}&5.2\\
&0.040&45.0&\textbf{5.0}&566.1&\textbf{47.5}&11.9\\
\bottomrule
\end{tabular}}
\label{table-relu1}
\end{table*}

\begin{table*}[!t]
    \caption{The performance of Ti-Lin and MaxPool2ReLU on ERAN framework.}
   \centering
   
       \resizebox{0.9\textwidth}{!}{
   \begin{tabular}{lrrrrrr}
   \toprule
   &\multicolumn{1}{l}{}&\multicolumn{2}{c}{{Certified accuracy(\%) }}&\multicolumn{2}{c}{{Averaged Time(s) }}&\multicolumn{1}{c}{{Speedup}}\\\cmidrule(lr){5-6}\cmidrule(lr){3-4}\cmidrule(lr){7-7}
   \multicolumn{1}{l}{{Network}}&\multicolumn{1}{l}{{Radius}}&\multicolumn{1}{c}{{MaxPool2ReLU}}&\multicolumn{1}{c}{{Ti-Lin}}&\multicolumn{1}{c}{{MaxPool2ReLU}}&\multicolumn{1}{c}{{Ti-Lin}}&\multicolumn{1}{c}{{vs. MaxPool2ReLU}}\\\midrule
\multirow{3}{*}{\shortstack[l]{MNIST\_Conv\_MaxPool}}
   &0.006&24.7&\textbf{69.1}&645.3&\textbf{18.5}&{34.9}\\
   &0.007&11.1&\textbf{64.2}&776.2&\textbf{25.7}&30.2\\
   &0.008&7.4&\textbf{45.7}&882.3&\textbf{36.9}&23.9\\\midrule
\multirow{3}{*}{\shortstack[l]{ConvNet\_MaxPool}}
   &0.020&0.0&\textbf{65.0}&984.3&\textbf{86.5}&{11.4}\\
   &0.030&0.0&\textbf{50.0}&1022.5&\textbf{153.5}&6.7\\
   &0.040&0.0&\textbf{35.0}&973.6&\textbf{149.6}&6.5\\
   \bottomrule
   \end{tabular}}
   \label{table-relu2}
   \end{table*}

\subsection{Experiment setup}

In this subsection, we present some experiment setups of Section~\ref{section5} in detail. 
Concretely, we list the number of nodes, the sources of networks, the number of Properties to be verified, and the perturbation range $\epsilon$ in Table~\ref{table-network}. Following the setting of ERAN, we generate properties for all networks by the correctly classified inputs in the first 100 inputs. We evaluate our method on four datasets, including MNIST, a dataset of $28\times28$ handwritten digital images in 10 classes, CIFAR-10,  a dataset of 60,000 $32\times32\times3$ images in 10 classes, e.g., airplane, bird, and ship, Tiny ImageNet~\cite{deng2009imagenet}, a dataset of 100,000  $64\times64\times3$  images in 200 classes, and ModelNet40, a dataset of 12,311 pre-aligned shapes from 40 categories. 

\subsection{Additional experiments}

In this subsection, we conduct some additional experiments to further illustrate (I) the performance of Ti-Lin on CNN-Cert, a back-substitution-based verifier.
(II) We compare Ti-Lin to OSIP to illustrate the superiority of tight linear approximation for MaxPool over the MaxPool2ReLU transformation.

\subsubsection{Results (I): performance on CNN-Cert}

To compare Ti-Lin to CNN-Cert fairly, we implement Ti-Lin on CNN-Cert. We follow the metrics used in CNN-Cert. We use the certified robustness bound introduced in Section~\ref{section3} as the tightness metric and the average computation time as the efficiency metric. As for the improvement of tightness, we use $\frac{100(\epsilon_{l}'-\epsilon_{l})}{\epsilon_{l}}\%$ to quantify the percentage of improvement, where $\epsilon_{l}'$ and $\epsilon_{l}$ represent the average certified lower bounds certified by Ti-Lin and CNN-Cert, respectively. We evaluate Ti-Lin and CNN0Cert on 10 inputs for all CNNs in Table~\ref{table-cnncert}. The initial perturbation range is 0.005. Testing on 10 inputs can sufficiently evaluate the performance of verification methods, as it is shown that the average certified results of 1000 inputs are similar to 10 images~\cite{boopathy2019cnn}. 
The results are shown in Table~\ref{table-cnncert}. 
The results indicates that, under different networks, datasets, perturbation norm($l_1,l_2,l_\infty$), Ti-Lin's certified bounds achieved better performance than CNN-Cert without incurring additional time overhead. Specifically, Ti-Lin exhibits larger robustness bounds compared to  CNN-Cert with improvements of up to  69.0, 43.3, 64.6\% on the MNIST, CIFAR-10, and Tiny ImageNet datasets, respectively.
\subsubsection{Results (II): comparison to MaxPool2ReLU transformation method}
\begin{figure*}[!t]
\centering
\includegraphics[width=1.0\columnwidth]{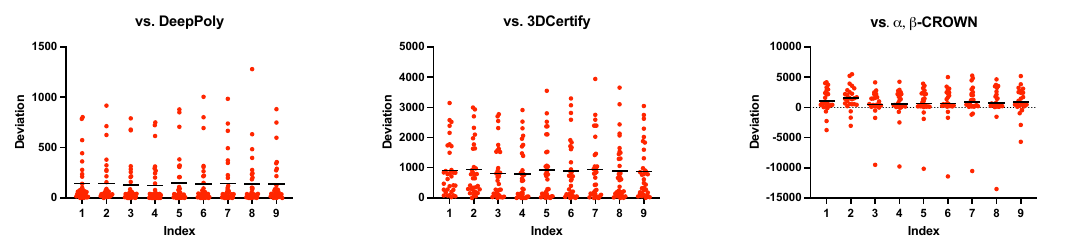} 
\caption{Visualization of the global lower bounds verified by DeepPoly, 3DCertify, $\alpha,\beta$-CROWN, and Ti-Lin. Red dots represent the deviation of the global bounds $\boldsymbol{L}-\boldsymbol{L'}$, where $\boldsymbol{L}$ and $\boldsymbol{L'}$ represent the global bounds of Ti-Lin and other methods testing on 100 inputs, respectively.  Black lines represent the mean of the deviations.}
\label{fig-raw}
\end{figure*}
 Notably, as OSIP~\cite{osip} is not open-sourced, we use another alternative method, called MaxPool2ReLU for evaluation. Concretely, we follow OSIP to transform the MaxPool-based network into a ReLU-based network.
  ERAN, employing advanced multi-ReLU relaxation for ReLU, and $\alpha,\beta$-CROWN, leveraging the branch-and-bound technique for ReLU neurons, are two cutting-edge verifiers designed for handling the ReLU layer. Therefore, we use the results of the transformed networks tested by ERAN and $\alpha,\beta$-CROWN as the alternative results of OSIP, denoted as MaxPool2ReLU. The results are shown in Table~\ref{table-relu1} and Table~\ref{table-relu2}. The results show that MaxPool2ReLU transformation would lead to coarse certified results and much more time consumption. Concretely, Ti-Lin computes higher certified accuracy with up to 35\% and 65\% improvement than MaxPool2ReLU on $\alpha,\beta$-CROWN and ERAN, respectively. Further, Ti-Lin can accelerate the verification process with up to 420.9$\times$ and 34.9$\times$ speedup regarding the average time of all instances on $\alpha,\beta$-CROWN and ERAN, respectively.
  This is because the transformation would make the network deeper, leading to the cumulative overapproximation region and much time consumption to verify. For example, when the pool size is $2\times 2$, one MaxPool layer can be transformed into three ReLU layers and three affine layers. 
  Thus, Ti-Lin is much faster and tighter than MaxPool2ReLU. Especially when verifying ConvNet\_MaxPool, the verified accuracy of MaxPool2ReLU (ERAN)  is zero across all perturbation radii, while Ti-Lin can at least verify 35\% inputs when the perturbation radius is 0.040, respectively.

\subsubsection{Result (III): Evaluation on global lower bounds}

According to Equation~\ref{eq-globallowerbound}, we decide whether the perturbed region is safe based on $l_t^K-u_j^K\geq 0,j\neq t, j\in [n_K]$. Therefore, the global lower bounds $\boldsymbol{L}:=(l_t^K-u_j^K),j\neq t, j\in [n_K]$ is the raw criterion to evaluate the tightness.
To further illustrate the advantages of the neuron-wise tightness over other linear bounds, we analyze the global lower bounds of the last layer computed by Ti-Lin and other methods, including DeepPoly, 3DCertify, and optimized linear bounds, used in $\alpha,\beta$-CROWN. 
 In Figure~\ref{fig-raw}, we show the deviation of the global lower bounds of Ti-Lin and the baseline on CIFAR\_Conv\_MaxPool. The $x$-axis represents the index of the global lower bound (labels without the true label), and the $y$-axis represents the deviations between the global lower bounds. 
As shown in Figure~\ref{fig-raw}, Ti-Lin has larger global bounds on all inputs than DeepPoly and 3DCertify and on most inputs than $\alpha,\beta$-CROWN. Further, the mean of the deviations $\boldsymbol{L}-\boldsymbol{L'}$ are all larger than zero. It reveals that the neuron-wise tightest linear bounds can bring tighter output bounds than other methods. Consequently, Ti-Lin can certify much larger certified accuracy than these methods in Table~\ref{table-3dcertify} and~\ref{table-alpha}.

\begin{figure*}[!t]
\centering
\includegraphics[width=0.4\columnwidth]{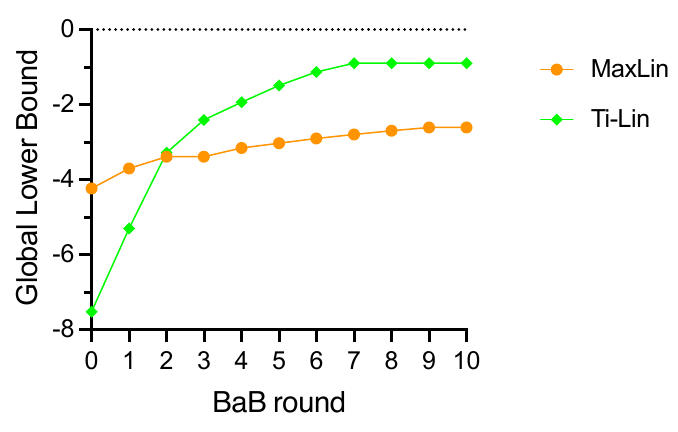} 
\caption{Visualization of the global lower bound verified by MaxLin and Ti-Lin, both of which are built upon the framework of $\alpha,\beta$-CROWN.}
\label{fig-maxlin}
\end{figure*}
According to the BaB design of $\alpha,\beta$-CROWN, we compare the global lower bound for verifying the property 0(true label agaist label 0) on ConvNet\_MaxPool, with the results illustrated in Figure~\ref{fig-maxlin}.
Initially, at BaB round = 0,  MaxLin  achieves a higher global lower bound compared to Ti-Lin. This is attributed to MaxLin’s block-wise tightest approach, which is specifically designed for ReLU + MaxPool blocks, giving it an early advantage in representing tighter bounds before the BaB analysis progresses.
However, as the BaB process progresses, the $\alpha,\beta$-CROWN verification framework employs plane-cutting techniques for ReLU neurons. This enables Ti-Lin’s neuron-wise tightest method to leverage its finer granularity, which significantly improves the global lower bound in later BaB rounds. Consequently, Ti-Lin surpasses MaxLin, achieving a much higher global lower bound in the later stages of analysis. This comparison demonstrates that while MaxLin provides better bounds in the initial stages, Ti-Lin’s more precise neuron-wise tightness proves to be advantageous for verifying robustness after iterative BaB analysis.

\subsubsection{Result (IV): Hybrid-Lin: a combination of Ti-Lin and MaxLin}

\begin{figure}[!t]
  \begin{minipage}[c]{0.45\textwidth}
    \centering
\centering
\includegraphics[width=\columnwidth]{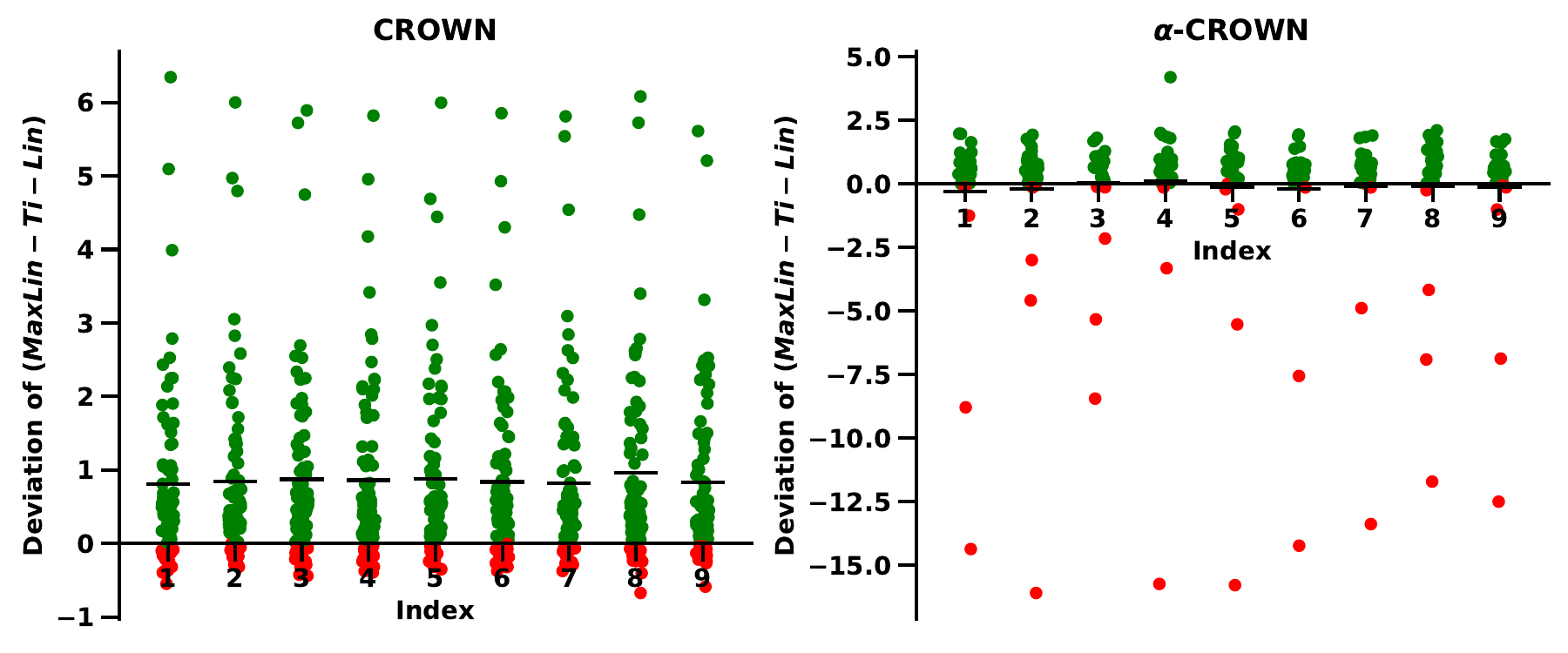} 
\captionsetup{skip=2pt}
\captionof{figure}{The deviation in global bounds between MaxLin and Ti-Lin when using the CROWN and $\alpha$-CROWN frameworks.}
\label{maxlin}
\vspace{-3mm}
  \end{minipage}
      \hfill
\centering
    \begin{minipage}[c]{0.45\textwidth}
\includegraphics[width=\columnwidth]{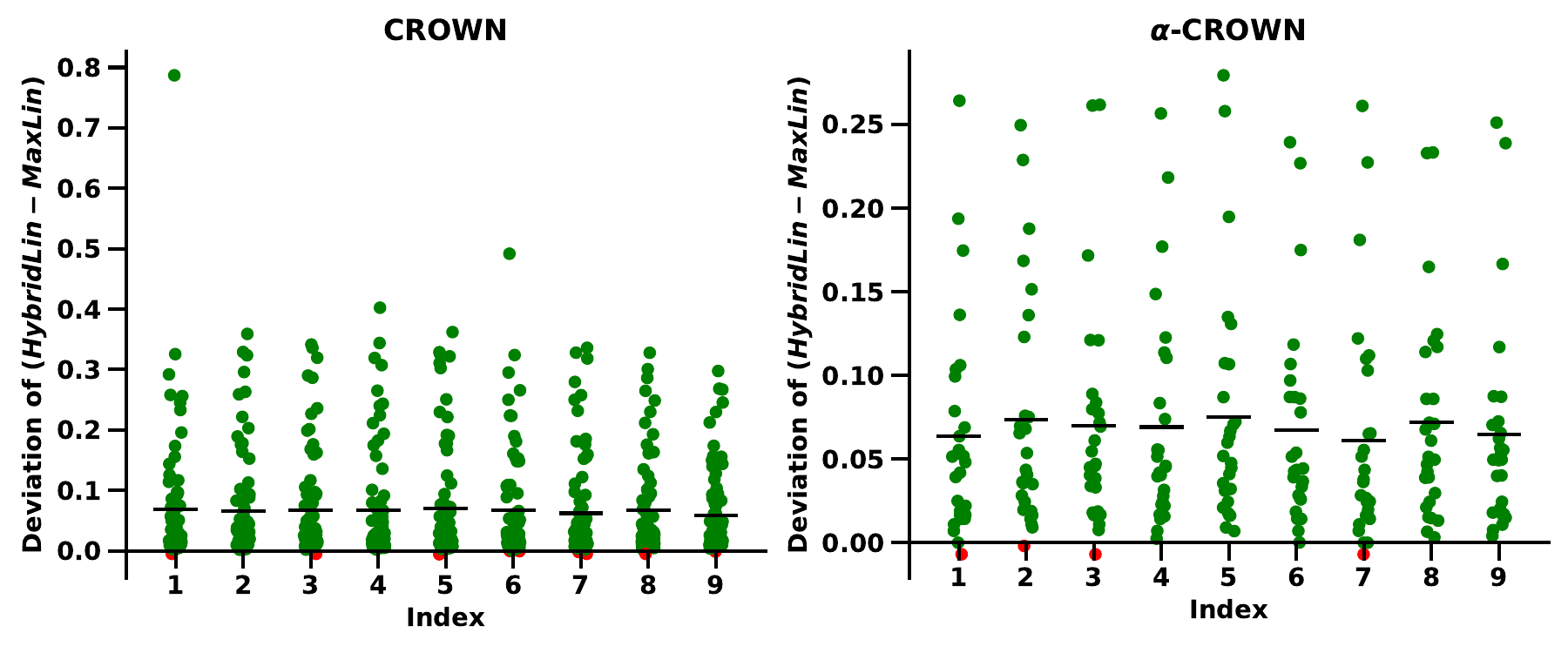} 
\captionsetup{skip=2pt}
\captionof{figure}{The deviation in global bounds between Hybrid-Lin and MaxLin when using the CROWN and $\alpha$-CROWN  frameworks.}
\label{hybridlin}
\vspace{-5mm}
  \end{minipage}
\end{figure}

{MaxLin and Ti-Lin make distinct but non-contradictory claims}. MaxLin provides the  \textbf{block-wise tightest} upper linear bound when the ReLU's upper linear bound is $u(x)=\frac{u}{u-l}(x-l)$, while Ti-Lin, being \textbf{neuron-wise tightest}, also achieves block-wise tightest bounds when the ReLU's upper linear bound is $u(x)=0$ or $u(x)=x$. 
Thus, Ti-Lin outperforms MaxLin when ReLU’s upper bound incurs no precision loss. 
Rather than competing, they complement each other in achieving optimal bounding precision.  To illustrate this, we introduce \textcolor{red}{HybridLin}, which uses MaxLin when one of the ReLU' upper linear bound is $u(x)=\frac{u}{u-l}(x-l)$ while Ti-Lin when all the ReLU's upper linear bound is $u(x)=0$ or $u(x)=x$.
In Figures~\ref{maxlin} and~\ref{hybridlin}, green points indicate positive deviation, red indicate negative.  
Figure~\ref{maxlin} shows that when ReLU’s upper linear bound incurring precision loss, MaxLin’s performance varies, underscoring the importance of Ti-Lin in achieving  tighter results.
Figure~\ref{hybridlin} shows that Hybrid-Lin, using Ti-Lin when ReLU’s upper linear bound is precise, consistently achieves tighter bounds than MaxLin.

\end{document}